\RequirePackage[svgnames]{xcolor}
\documentclass[11pt,letterpaper,onecolumn]{mystyle}

\usepackage[table]{xcolor}
\usepackage{url}
\usepackage{xspace}
\usepackage{graphicx}
\usepackage{subcaption}
\usepackage{listings}
\usepackage{xcolor}
\usepackage{titletoc}
\usepackage{amsmath}
\usepackage[font={small}]{caption}
\usepackage[export]{adjustbox}
\usepackage{bm}
\usepackage{booktabs}
\usepackage{balance}
\usepackage{color}
\usepackage{dsfont}
\usepackage{esvect}
\usepackage{float}
\usepackage{graphicx}
\usepackage{pifont}
\usepackage{import}
\usepackage{mathtools}
\usepackage{multirow}
\usepackage{stackengine}
\usepackage{stmaryrd}
\usepackage{systeme}
\usepackage{soul}
\usepackage{tabularx}
\usepackage{url}
\usepackage{tikz}
\usetikzlibrary{tikzmark}
\usetikzlibrary{calc}
\usepackage[normalem]{ulem}
\usepackage{enumitem}
\usepackage{algorithm2e}
\usepackage{scalerel}
\usepackage{wrapfig}
\usepackage{bxcoloremoji}

\SetKwComment{Comment}{\color{green!50!black}\# }{}

\SetKwProg{Function}{def}{:}{}

\SetKwProg{For}{for}{:}{}
\SetKwProg{If}{if}{:}{}

\usepackage{booktabs}    % \toprule \midrule \bottomrule
\usepackage{makecell}    % \makecell
\usepackage{arydshln}    % \hdashline
\usepackage{colortbl}    % \arrayrulecolor
\usepackage{xcolor}      % \textcolor (and custom colors)
\usepackage{stfloats}
\usepackage{nicefrac}

\definecolor{turquoise}{cmyk}{0.65,0,0.1,0.3}
\definecolor{purple}{rgb}{0.65,0,0.65}
\definecolor{dark_green}{rgb}{0, 0.5, 0}
\definecolor{orange}{rgb}{0.8, 0.6, 0.2}
\definecolor{red}{rgb}{0.8, 0.2, 0.2}
\definecolor{darkred}{rgb}{0.6, 0.1, 0.05}
\definecolor{blueish}{rgb}{0.0, 0.3, .6}
\definecolor{light_gray}{rgb}{0.7, 0.7, .7}
\definecolor{pink}{rgb}{1, 0, 1}
\definecolor{greyblue}{rgb}{0.25, 0.25, 1}
\definecolor{LightRed}{rgb}{0.99,0.89,0.89}
\definecolor{copecolor}{HTML}{4E95D9}

\newcommand{\xmark}{\ding{55}} 

\newcommand{\project}{CoPE-VideoLM}

\definecolor{colorFst}{HTML}{C8E6C9}   % 1st place - Soft Green
\definecolor{colorSnd}{HTML}{FFF9C4}   % 2nd place - Light Yellow/Amber
\definecolor{colorTrd}{HTML}{FFCDD2}   % 3rd place - Light Red/Salmon

\newcommand{\fs}{\cellcolor{colorFst}\bf}
\newcommand{\nd}{\cellcolor{colorSnd}}
\newcommand{\rd}{\cellcolor{colorTrd}}

\definecolor{teaser_gray}{rgb}{0.88, 0.89, 0.90} 
\definecolor{teaser_green}{rgb}{0.77, 0.88, 0.70} 
\definecolor{teaser_blue}{rgb}{0.70, 0.78, 0.90} 
\definecolor{teaser_yellow}{rgb}{1.0, 0.90, 0.60} 
\definecolor{codegray}{rgb}{0.7,0.7,0.7}
\definecolor{motiongreen}{HTML}{4EA72E}
\definecolor{residualpink}{HTML}{A02B93}

\newcommand{\PAR}[1]{\vskip4pt \noindent{\bf #1~}}

\usepackage{enumitem}
\setlist{topsep=2pt,itemsep=1pt,parsep=0pt,partopsep=0pt}

\definecolor{citecolor}{HTML}{0071bc}
\usepackage{hyperref}[citecolor=citecolor]

\title{\project: Leveraging Codec Primitives For Efficient Video Language Modeling}

\author[1,2 \circ]{Sayan Deb Sarkar}
\author[1]{Rémi Pautrat}
\author[1]{Ondrej Miksik}
\author[1,3]{Marc Pollefeys}
\author[2]{Iro Armeni}
\author[1^*]{Mahdi Rad}
\author[1^*]{Mihai Dusmanu}

\affiliation[1]{Microsoft Spatial AI Lab}
\affiliation[2]{Stanford University}
\affiliation[3]{ETH Zurich}

\date{March 30, 2026}
\correspondence{Sayan Deb Sarkar at \email{sdsarkar@stanford.edu}}

\contribution[\circ]{Part of work done at Microsoft}
\contribution[*]{Equal Supervision}

\abstract{
Video Language Models (VideoLMs) enable AI systems to understand temporal dynamics in videos. To fit within the maximum context window constraint, current methods use keyframe sampling which often misses both macro-level events and micro-level details due to the sparse temporal coverage. Furthermore, processing full images and their tokens for each frame incurs substantial computational overhead. We address these limitations by leveraging video codec primitives (specifically motion vectors and residuals) which natively encode video redundancy and sparsity without requiring expensive full-image encoding for most frames. To this end, we introduce lightweight transformer-based encoders that aggregate codec primitives and align their representations with image encoder embeddings through a pre-training strategy  that accelerates convergence during end-to-end fine-tuning. Our approach, \textit{\project}, reduces the time-to-first-token by up to $86\%$ and token usage by up to $93\%$ compared to standard VideoLMs. Moreover, by varying the keyframe and codec primitive densities we maintain or exceed performance on $14$ diverse video understanding benchmarks spanning general question answering, temporal and motion reasoning, long-form understanding, and spatial scene understanding.
}

\metadata[\coloremojicode{1F3E0} \textbf{Project}]{\href{https://microsoft.github.io/CoPE}{https://microsoft.github.io/CoPE}}

\teaser{%
  \begin{center}
    \vspace{4pt}
    \includegraphics[width=\textwidth]{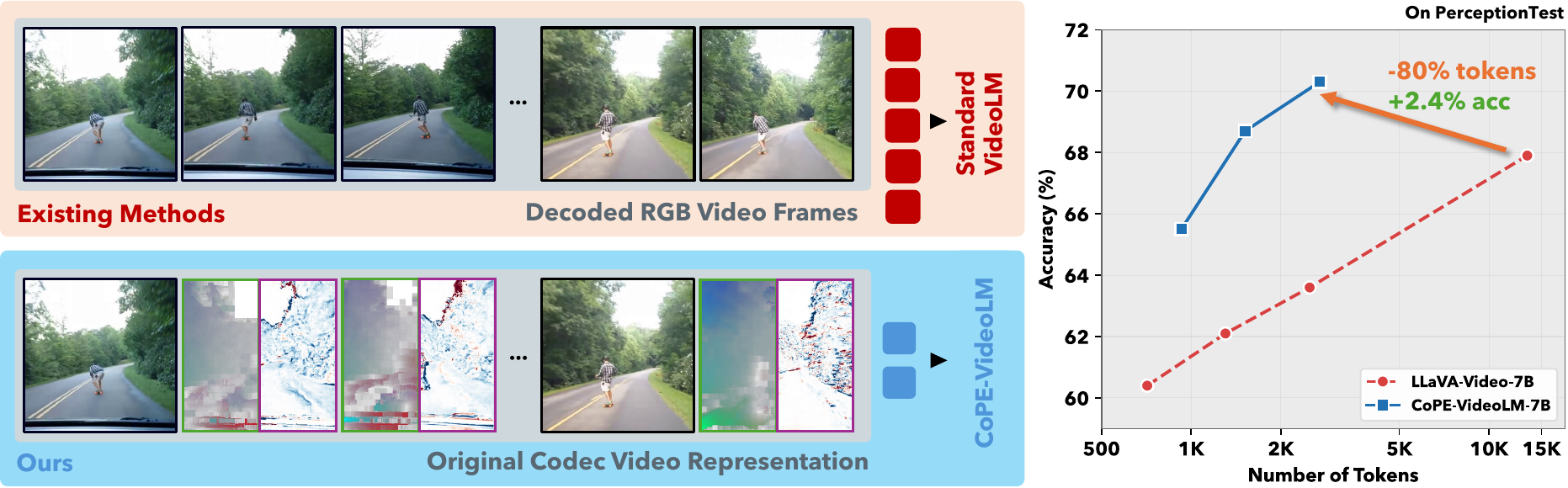}
    \captionof{figure}{\textbf{\textit{\project{}}} is a codec-aware tokenization framework for Video Language Models that replaces dense RGB frame encoding with lightweight structured representations derived from codec primitives. Instead of treating every frame as a full image, the model processes only sparse I-frames through a vision encoder, while P-frames are converted into compact tokens using their \textcolor{motiongreen}{motion vectors} and \textcolor{residualpink}{residuals}. By leveraging the inherent sparsity and structure of standard video codecs, CoPE-VideoLM avoids redundant RGB processing, reduces visual token usage by up to $93\%$, and cuts time-to-first-token by up to $86\%$, compared to standard dense VideoLMs.}
    \label{fig:teaser}
  \end{center}
}

\begin{document}
\maketitle

\section{Introduction}
\label{sec:intro}

Video Language Models (VideoLMs) represent a major advancement in multi-modal AI~\cite{2023videochat,lin2023video,zhu2023languagebind,zhang2024videoinstructiontuningsynthetic,damonlpsg2025videollama3}, enhancing Vision Language Models (VLMs) with temporal reasoning that allows them to understand how visual narratives, objects, actions, and relationships evolve across video sequences.
This enables a wide range of downstream applications, from more natural human-computer interaction~\cite{humanllm} through video question-answering to robotics~\cite{kawaharazuka2025vlasurvey,ma2025surveyvisionlanguageactionmodelsembodied}, where agents must understand sequential actions~\cite{liu2025trivlatriplesystembasedunifiedvisionlanguageaction,yang2025egovlalearningvisionlanguageactionmodels}.
Beyond these applications, VideoLMs represent a step toward AI systems that process visual information not as disconnected snapshots but as continuous, temporally coherent experiences.
As video continues to dominate digital content, models that can ``watch'' and understand video at scale~\cite{Achiam2023GPT4TR,comanici2025gemini,Qwen2.5-VL} become essential for the next generation of AI applications.

However, realizing this vision remains challenging: VideoLMs have a maximum context window limiting the amount of information that can be provided as input.
This is not only a training artifact, but also related to hardware constraints, as larger context windows require linearly more memory and quadratically more compute~\cite{shah2024flashattention}.
To fit in the context window, existing VideoLMs select a subset of video frames as keyframes either through hand-crafted heuristics (e.g., uniform temporal sampling~\cite{reid2024gemini}) or learned methods (e.g., flexible frame selection~\cite{buch2025flexible} or token compression~\cite{llavascissor}).
In this setup, proprietary models with extended context lengths of up to 1 million tokens can process one hour of video at 1 FPS~\cite{reid2024gemini}.
Open-source models~\cite{zhang2024videoinstructiontuningsynthetic} have a much smaller budget and sample a fixed number of frames (e.g., $64$) regardless of video length. This is a severe limitation, as the information content naturally scales with video duration.

Moreover, any keyframe sampling approach fundamentally limits VideoLMs' understanding by providing only sparse temporal coverage: this can miss both macro-level events crucial for high-level comprehension and micro-level details necessary for recognizing fine-grained actions~\cite{slowfocus,longvu,temporalquery}.
Compounding this, there is generally high redundancy between consecutive frames (even when downsampled to 1 FPS); therefore, using the same token budget for each keyframe is suboptimal. 
Furthermore, treating each keyframe as a full image also significantly increases the prefill time~\cite{vasu2025fastvlm}, delaying the \textit{time-to-first token} (TTFT) of the VideoLM.
Low TTFT is critical for user experience and essential for robotics applications requiring real-time responsiveness.

A natural source of structure that addresses both redundancy and sparsity is already present in video processing pipelines: video codecs, a decades-old field designed precisely to solve these problems~\cite{mpeg,Wiegand2003OverviewOT}.
Rather than encoding every frame as a full image, codecs encode what moves between frames as motion vectors and the intensity changes as residuals, preserving temporal structure while minimizing redundancy.
In a typical streaming setup, codecs select keyframes (I-frames) every 5--10 seconds, either at fixed intervals or when large scene changes are detected, encoding only the changes for all remaining frames (P-frames).

Prior works have explored codec-domain signals, but with important limitations: CoViAR~\cite{wu2018coviar} and TeamNet~\cite{teamnet} trained separate CNNs on motion vectors and residuals, ignoring cross-modal dependencies for action recognition. Video-LaVIT~\cite{videolavit} discretizes motion vectors but discards residuals. EMA~\cite{ema2025} aggregates the I-frame and motion vectors from a group of pictures (GOP) into a fixed-length summary, also discarding residuals and losing temporal ordering.
Critically, no existing method preserves both dynamics and appearance in a variable-length, temporally ordered representation.

We propose to directly encode motion vectors and residuals (i.e., codec primitives) as temporally aligned tokens compatible with image features, and arrange them in temporal order alongside keyframe tokens to form a codec-aware token sequence during inference. An overview of our approach is shown in Fig.~\ref{fig:teaser}.
This provides two major advantages: first, we avoid the costly full image encoding for most frames, and second, we can use far fewer tokens given the sparse nature of these primitives, thus reducing TTFT by a significant margin.
Optionally, the motion vectors and residuals of several subsequent frames can be grouped together to strike a trade-off between fine-grained representation and total number of tokens.
While we focus on VideoLMs, the methodology is in principle applicable to other tasks such as video retrieval or action recognition.
Our formulation addresses all three gaps identified above: we preserve residuals, maintain temporal ordering through interleaved I/P-frame tokens, and allow the token budget to scale with information content.

To this end, we use transformer-based encoders to aggregate the information of all motion vectors and residuals from a given GOP. 
These encoders are first pre-trained to adapt them to the space of the image encoder and then integrated with a VideoLM for end-to-end fine-tuning. Our model matches the performance of current open-weight VideoLMs and even surpasses them on several benchmarks despite using substantially fewer tokens.
We extensively validate our approach across $14$ benchmarks, demonstrating consistent gains across general video QA, temporal and motion reasoning, long-form reasoning, and spatial scene understanding tasks.
Our main contributions are as follows:
\begin{itemize}
\item We propose to encode videos for VideoLMs by leveraging codec primitives, preserving motion and appearance in temporal order. Codec primitives allow us to skip redundant RGB information, reducing the TTFT by up to $86$\% and token usage by up to $93$\%.
\item We introduce a lightweight dual-branch architecture for encoding codec primitives, achieving substantially higher compression rates and lower token counts than traditional image-based encoders.
\item We propose a pre-training framework for codec-primitive encoders that aligns their representation space with that of image encoders, enabling faster training and quicker convergence when integrated with the VideoLM.
\end{itemize}
\section{Related Work}
\label{sec:related_work}

\PAR{Video Language Models.}
Recent advances in Multimodal Large Language Models (MLLMs)~\cite{videounderstandingsurvey,largescalemultimodalsurvey,liu2023llava,liu2023improvedllava,liu2024llavanext,li2024llava,zhang2024longva,internlmxcomposer,internlmxcomposer2_5,timemarker,chen2024internvl2.5,apollo} have extended image-based architectures into the video domain, giving rise to VideoLMs capable of temporal reasoning over dynamic visual content. MLLMs typically comprise a vision encoder (e.g., CLIP~\cite{clip} or SigLIP~\cite{siglip}), a modality alignment mechanism also known as adapter (e.g., linear projection, Q-Former~\cite{zhang2022vsa,zhang2023vision}, or gated cross-attention~\cite{flamingo}), and an LLM backbone (e.g., LLaMA~\cite{llama}, Vicuna~\cite{vicuna}, or Qwen~\cite{Qwen-VL,Qwen2-VL,Qwen2.5-VL}) for multimodal decoding. Early VideoLMs such as Video-LLaMA~\cite{damonlpsg2023videollama} and VideoChat2~\cite{2023videochat} were limited by short context and redundant tokenization. Subsequent models improve efficiency through extended context windows~\cite{damonlpsg2025videollama3}, token pooling or merging~\cite{Maaz2023VideoChatGPT,huang2024lita,qian2024momentor,wang2024hawkeye}, timestamp-aware encoding~\cite{timechat}, and large-scale instruction tuning~\cite{zhang2024videoinstructiontuningsynthetic}. Closed-source systems like Gemini~\cite{reid2024gemini,gpt5,comanici2025gemini}, GPT~\cite{gpt5,Achiam2023GPT4TR} and Claude~\cite{anthropic2025sonnet} demonstrate impressive fine-grained and long-context understanding but depend on proprietary data and undisclosed architectures. Despite these advances, open-source VideoLMs still process videos as dense RGB frame collections, overlooking the structured redundancy inherent in standard video codecs. Leveraging codec primitives like motion vectors and residuals, our approach directly encodes temporal dynamics, supporting efficient long-context understanding while retaining fine-grained detail.

\PAR{Token Compression.} Compression techniques reduce the number of visual tokens fed into VideoLMs by removing redundancy while preserving semantic fidelity. Existing methods can be grouped into \emph{heuristic} and \emph{learnable} approaches. Heuristic methods apply uniform downsampling~\cite{lin2023vila}, spatial or temporal pooling~\cite{xu2024pllava,zhang2024videoinstructiontuningsynthetic,li2024llamavid}, or similarity-guided merging~\cite{bolya2022tome,jin2023chatunivi,li2024videochat}. Learnable modules such as Q-Former~\cite{liblip2,Qwen-VL,llavamini}, Perceiver Resampler~\cite{yao2024minicpm}, and memory-based mechanisms~\cite{song2023moviechat,song2024moviechat+,jin2023chatunivi} generate compact latent representations before passing them to the LLM. Attention-based methods~\cite{fastv,fu2024framefusion,Xing2024PyramidDropAY,zhang2024sparsevlm} exploit the observation that visual tokens receive diminishing attention in deeper layers~\cite{fastv,shao2025tokens}, pruning or modulating token ratios across layers to remove redundancy. Temporal pooling approaches~\cite{xu2024pllava,Maaz2023VideoChatGPT,slowfastllava,longvlm} exploit inter-frame redundancy by downsampling at the frame level, while DyCoke~\cite{dycoke} and LLaVA-Scissor~\cite{llavascissor} further leverage spatio-temporal structure for compression (see Supp. for a direct comparative study; codec-native sparsity provides a stronger starting point than post-hoc pruning). More adaptive approaches such as AdaReTake~\cite{adaretake} and FlexSelect~\cite{flexselect} dynamically allocate compression budgets across layers or leverage cross-modal attention to filter tokens without retraining. However, these methods rely on dense RGB frame encodings and neglect intrinsic temporal redundancy. In contrast, our native codec representation inherently encodes only meaningful temporal changes rather than removing information post hoc.

\PAR{Compressed Video Representation.} There has been growing interest in exploiting motion vectors and residuals directly from compressed video streams for visual understanding, particularly in action recognition, thus bypassing the costly full-frame processing. Early works such as CoViAR~\cite{wu2018coviar} and TEAM-Net~\cite{teamnet} trained separate CNNs on I- and P-frame signals but ignored inter-modal dependencies and temporal ordering, requiring costly multi-clip inference. Later methods extended compressed-domain learning to 3D CNNs~\cite{cvc3d}, optical-flow distillation~\cite{dmcnet,havasi2021training,compressedteacher}, and transformer-based self-attention~\cite{Chen2021MMViTMV}, improving accuracy but still incurring high inference costs or requiring decoded RGB frames during training. CompressedVideoMAE~\cite{biswas2025scalablemodelingcompressedvideos} demonstrates that masked autoencoding on motion vectors and residuals alone can match raw-video pretraining at far less compute. More recently, codec-based representations have been explored in VideoLMs. Video-LaVIT~\cite{videolavit} discretizes motion vectors into language-like tokens, and EMA~\cite{ema2025} aggregates I-frames and motion vectors into a fixed-length GOP summary, similar to Video-VAE~\cite{videovae}. These approaches either discard residuals or collapse temporal ordering; {\project} addresses both by constructing a variable-length, temporally ordered token sequence that preserves fine-grained motion and appearance signals.
\begin{figure}[t]
    \centering
    \includegraphics[width=\linewidth]{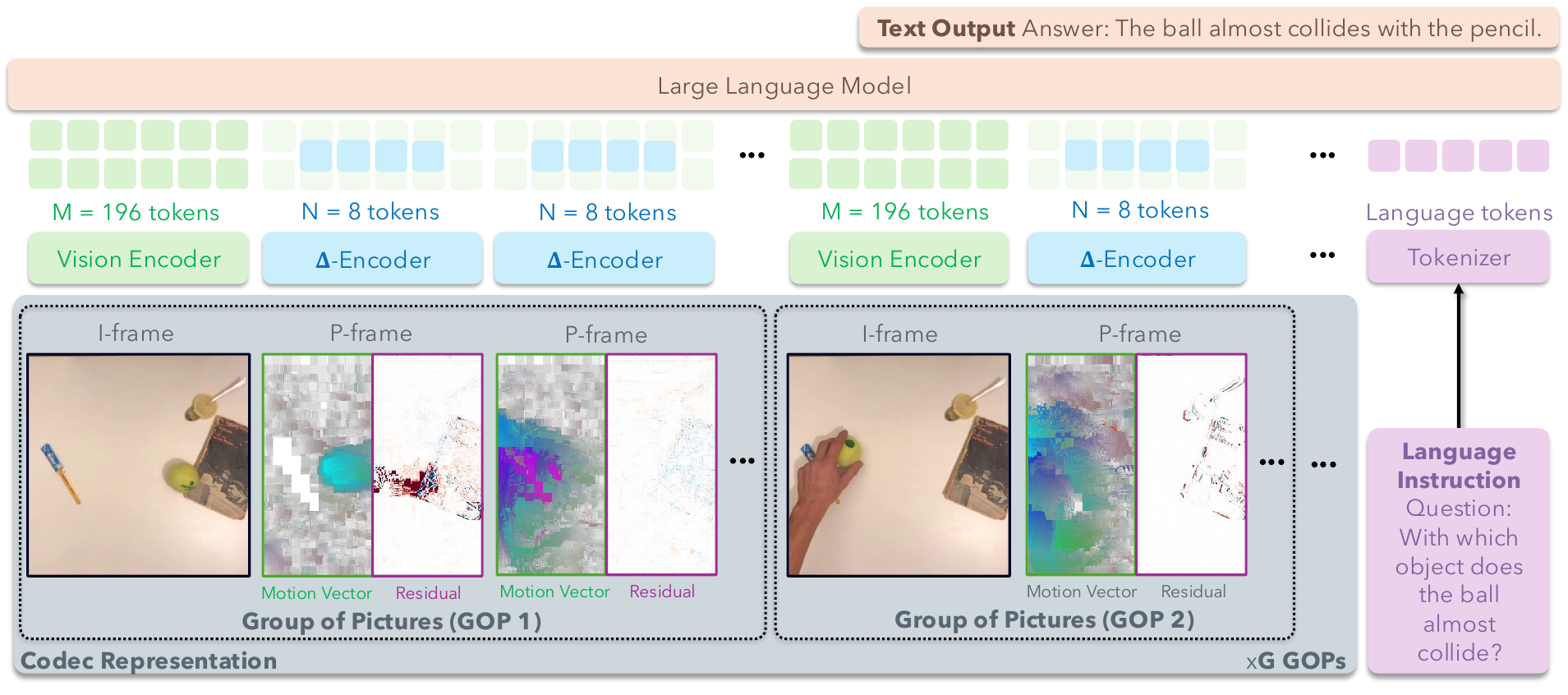}
    \caption{
    \textbf{Overview of our pipeline}. Given a video in its raw codec representation, our framework leverages the GOP structure for efficient, codec-aware tokenization. \textit{I-frames} are processed by a standard frozen vision encoder ($\phi_{\text{RGB}}$) to produce dense RGB tokens. \textit{P-frames}, however, bypass full RGB decoding. Their raw components, \textcolor{motiongreen}{motion vectors} and \textcolor{residualpink}{residuals}, are instead fed into our lightweight $\Delta$-Encoder ($\phi_{\Delta}$) to generate a small set of highly compact $\Delta$-tokens. The final token stream, an interleaved sequence of I-frame tokens and $\Delta$-tokens, is consumed by the LLM, enabling dense temporal coverage at a fraction of the standard token count and runtime.
    }
\label{fig:pipeline}
\end{figure}

\section{Method}
\label{sec:method}

\subsection{Preliminaries}
\label{sec:preliminaries}

Modern video codecs such as MPEG-4, H.264, and HEVC~\cite{Grois2021PerformanceCO,Laude_Adhisantoso_Voges_Munderloh_Ostermann_2019} achieve high compression ratios by exploiting temporal redundancy across consecutive frames.
Let a video $\mathcal{V}$ be a sequence of frames $(F^{(1)}, \dots, F^{(T)})$.
Each of these frames can be an \textbf{I-frame}~(intra-coded), a \textbf{P-frame}~(predictive), or optionally a \textbf{B-frame}~(bi-directional predictive). Consecutive frames are organized in a \emph{Group of Pictures}~(GOP) structure as illustrated in Fig.~\ref{fig:pipeline}.

\PAR{I-frames.} An I-frame, $I^{(t)}$, is an RGB image encoded independently without the use of preceding or subsequent frames.
It is used as a reference point of the Group of Pictures and provides a full visual representation.

\PAR{P-frames.}
A P-frame \( P^{(t)} \) contains only the changes from the previous frame, be it a reference frame \( I^{(t-1)} \) or a preceding P-frame \( P^{(t-1)} \).
The difference is defined by two components:
\begin{itemize}
    \item \emph{Motion vectors} $\tau^{(t)}$, which describe block-wise displacements from the reference to the target frame, resembling coarse optical flow.  
    \item \emph{Residuals} $\delta^{(t)}$, which capture block-wise pixel corrections that remain after motion compensation.  
\end{itemize}
The recurrence~\cite{wu2018coviar,biswas2025scalablemodelingcompressedvideos} for reconstructing a P-frame is:
\begin{equation}
\hat{I}^{(t)}_i = \hat{I}^{(t-1)}_{i - \tau^{(t)}_i} + \delta^{(t)}_i \enspace ,
\label{eq:codec-eq}
\end{equation}
where $i$ is the pixel coordinate index. $\hat{I}^{(t-1)}$ is the reference frame at time $t{-}1$: it equals the raw I-frame $I^{(t-1)}$ when $F^{(t-1)}$ is intra-coded, and the reconstructed frame from earlier P-frames otherwise. Hence, P-frames contain only incremental temporal information, and are therefore smaller in file size than I-frames.

\PAR{B-frames.} A B-frame leverages both preceding and subsequent frames to encode its differences. While this bidirectional prediction achieves the highest compression efficiency, it increases decoding complexity: B-frames must wait for future I/P-frames, creating a mismatch between decode and display order. Consequently, they are less suited for streaming or real-time use. We therefore focus on P-frames, which depend only on past references and align naturally with the causal processing required by VideoLMs.

\PAR{Group of Pictures (GOP).} A GOP is a cycle structure that comprises one I-frame together with any mixture of P- and/or B-frames, e.g., $I\,B\,P\,P\,B\,P\,$etc. The GOP structure and length control the trade-offs between the compression efficiency, quality, and the capability for random access. Typical applications use varying configurations: all three frame types are used by the H.264 codec, while MPEG-4 uses the I- and P-frames (see Supp. for a detailed explanation).

\PAR{Implication for VideoLMs.} Despite this rich structure, current VideoLMs discard codec information and fully decode and tokenize dense RGB frames, ignoring the inherent sparsity of P- and B-frames. This results in unnecessary computation and inflated token counts, as shown in Sec.~\ref{sec:main_results}.

\subsection{\project{}}
 Our method explicitly leverages the GOP structure and introduces a \emph{codec-aware tokenization framework} that seamlessly integrates with VideoLMs.  Instead of unnecessarily encoding each frame as RGB patches~\cite{zhang2024videoinstructiontuningsynthetic} or compressing an entire GOP into a fixed-length summary~\cite{ema2025}, we retain I-frames as full RGB tokens and encode P-frames into lightweight and compact \textit{$\Delta$-tokens} (delta for difference) obtained from motion vectors and residuals. This preserves temporal ordering and appearance information without exceeding token budgets. An illustration of this process is presented in Fig.~\ref{fig:pipeline}.

Given a video $\mathcal{V} = (F^{(1)}, \dots, F^{(T)})$, each frame $F^{(t)}$ is represented as:
\begin{equation}
F^{(t)} =
\begin{cases}
I^{(t)}, & \text{if $F^{(t)}$ is an I-frame}, \\
P^{(t)} = (\tau^{(t)}, \delta^{(t)}), & \text{if $F^{(t)}$ is a P-frame},
\end{cases}
\label{eq:i_p_frame}
\end{equation}
where $\tau^{(t)}$ is a set of block-wise motion vectors (usually one per up to $16 \times 16$ macroblock), and $\delta^{(t)}$ is a set of block-wise residuals.
For our downstream processing, we represent these as tensors: $\tau^{(t)} \in \mathbb{Z}^{H \times W \times 2}$ is the sparse tensor constructed from the block-wise motion vectors and $\delta^{(t)} \in \mathbb{R}^{H \times W \times C}$ is the sparse tensor of residuals.

\PAR{I-frame processing.}  
I-frames $I^{(t)}$ are passed through a frozen vision encoder $\phi_{\text{RGB}}$, producing $M$ dense tokens:
\begin{equation}
X^{(t)}_{I} = \phi_{\text{RGB}}(I^{(t)}) \in \mathbb{R}^{M \times d}.
\end{equation}

\PAR{P-frame processing.}  
Each P-frame $P^{(t)}$ is mapped into a much more compact representation consisting of $N \ll M$ tokens by the $\Delta$-Encoder $\phi_{\Delta}$:
\begin{equation}
X^{(t)}_P = \phi_{\Delta}(\tau^{(t)}, \delta^{(t)}) \in \mathbb{R}^{N \times d}.
\end{equation}

\PAR{P-frame fusion.}
In the setup described so far, processing all frames at the native frame rate is mandatory as codec primitives are defined relative to previous frames and skipping frames invalidates these dependencies.
For example, let us consider a $30$ FPS video with a GOP size of $240$ frames ($8$ seconds). Our technique yields $M + 239N$ tokens per GOP, compared to $240M$ if all frames were encoded as RGB images.
Fine-grained action recognition may require this exhaustive temporal coverage, but most video understanding tasks can often be done with sparser coverage.

Rather than processing all frames at the native frame rate, we can fuse $s$ consecutive P-frames, encoding their combined changes relative to frame $F^{(t-s)}$ (rather than the immediately preceding frame).
The maximum number of P-frames that can be fused is bounded by the GOP size. In the running example, using $s=30$ P-frames (1 FPS) for fusion reduces the per-GOP token count to $M + 7N$, which is much smaller than $M + 239N$ for full P-frame modeling and  $8M$ for RGB encoding at $1$ FPS.
This fusion offers a codec-native way to trade temporal resolution for efficiency and can be tuned to match available compute and task requirements.

For clarity, we treat the temporal index $(t)$ as already incorporating the P-frame fusion, so $F^{(t)}$ always depends on $F^{(t - 1)}$, though $(t)$ may no longer correspond to the raw frame indices.

\begin{wrapfigure}{r}{0.52\textwidth}
    \centering
    \includegraphics[width=0.52\textwidth, trim=0 0 0 20, clip]{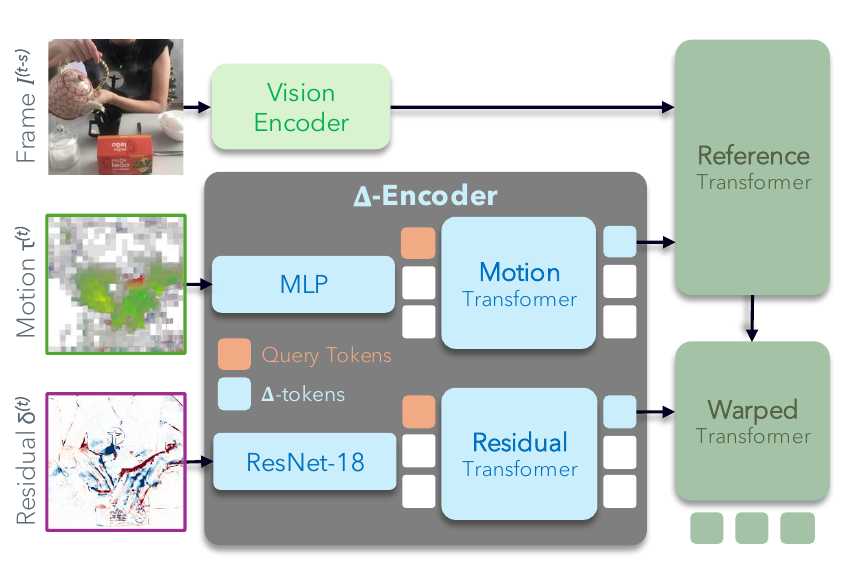}
    \caption{
    \small
    \textbf{$\Delta$-encoder} processes \textcolor{motiongreen}{motion vectors} and \textcolor{residualpink}{residuals} through two lightweight branches designed to extract and compress codec information. The resulting motion and residual tokens are concatenated to form the $\Delta$-tokens used for our efficient P-frame representation, which is projected to the RGB token space during pre-training.}
    \label{fig:delta_encoder}
\end{wrapfigure}

\PAR{$\Delta$-Encoder.}
The $\Delta$-Encoder $\phi_{\Delta}$ (Fig.~\ref{fig:delta_encoder}) is designed to process the motion vectors $\tau^{(t)}$ and residuals $\delta^{(t)}$ through two specialized branches.
The motion vectors $\tau^{(t)}$ are processed via a multi-layer MLP to extract local features over a grid of size $H_G \times W_G$ yielding features in $\mathbb{R}^{(H_G W_G) \times d}$. These features are then compressed via a motion transformer $\theta_{\texttt{motion}}$ with a set of $K_\tau$ learnable query tokens that can attend to all $H_G W_G$ input tokens and aggregate their information as:
\begin{equation}
    \tau^{(t)}_{\texttt{tok}} = \theta_{\texttt{motion}}(\texttt{MLP}(\tau^{(t)})) \enspace .
\end{equation}
Only these $K_\tau$ compressed motion tokens $\tau^{(t)}_{\texttt{tok}} \in \mathbb{R}^{K_\tau \times d}$ are used in the VideoLM.

The residual frame $\delta^{(t)}$ is embedded by a lightweight ResNet-18~\cite{He2015DeepRL} module to extract local features over the same grid size $H_G \times W_G$. Similarly to above, a residual transformer $\theta_{\texttt{residual}}$ aggregates and compresses the raw features to a set of $K_\delta$ compressed residual tokens $\delta^{(t)}_{\texttt{tok}} \in \mathbb{R}^{K_\delta \times d}$ as:
\begin{equation}
    \delta^{(t)}_{\texttt{tok}} = \theta_{\texttt{residual}}(\texttt{ResNet}(\delta^{(t)})) \enspace .
\end{equation}

% \PAR{Interleaved token stream.} In practice, we set $K_\tau = K_\delta = 4$ and thus, $N=8$ (see Supp. for an ablation study). The final visual sequence input to the LLM is an ordered concatenation of I-frame and P-frame tokens:

% \begin{equation}
% \label{eq:interleave}
% \begin{aligned}
% X &= \left[\, x^{(1)}, x^{(2)}, \dots, x^{(T)} \,\right], \\
% \text{where} \enspace x^{(t)} & =
% \begin{cases}
% X^{(t)}_I, & \text{if $F^{(t)}$ \text{ is an I-frame}} \\
% X^{(t)}_P, & \text{if $F^{(t)}$ \text{ is a P-frame}}
% \end{cases}
% \enspace .
% \end{aligned}
% \end{equation}

\PAR{Interleaved token stream.} In practice, we set $K_\tau = K_\delta = 4$ and thus $N=8$ (see Supp.; performance plateaus beyond $N=8$). The final visual sequence $X = [x^{(1)}, \dots, x^{(T)}]$ is formed by interleaving I-frame tokens $X^{(t)}_I$ and P-frame tokens $X^{(t)}_P$ in their natural temporal order, as shown in Eq.~\ref{eq:i_p_frame}. The LLM can consume $X^{(t)}_{I}$ and $X^{(t)}_{P}$ alongside textual instructions, without any architectural modifications. This approach reduces the redundant data considerably while preserving accurate temporal coverage and enables smooth scalability to long videos by adjusting the I-frame density, P-frame grouping, and token allocation.

\subsection{Training Paradigm}
\label{sec:training_paradigm}

The training is done in two stages.
First the $\Delta$-encoder is pre-trained in order to render it compatible with the image encoder.
Then, this encoder is integrated into a VideoLM and the whole pipeline is fine-tuned end-to-end.

\PAR{$\Delta$-encoder pre-training.}
The primary difficulty lies in ensuring that the $\Delta$-tokens \(X^{(t)}_P\) are aligned with the image tokens \(X^{(t)}_I\).
To achieve this, we pre-train the $\Delta$-Encoder as a modality adapter that enables codec-derived primitives \((\tau^{(t)}, \delta^{(t)})\) to be compatible with the embedding space defined by a vision encoder. By aligning the $\Delta$-tokens with this space, P-frames can then be compactly represented and replace standard RGB tokens within a VideoLM. For pre-training, two additional modules are used on top of the outputs of the $\Delta$-encoder.
First, a ``reference'' transformer $\theta_{\texttt{ref}}$ uses the image tokens from $I^{(t - 1)}$ and the compressed motion vector tokens $\tau^{(t)}_{\text{tok}}$, and its aim is to understand how the information moved in the image and how this displacement transforms the reference tokens.
This is akin to the warping in Eq.~\ref{eq:codec-eq}.
Second, a ``warped'' transformer $\theta_{\texttt{warped}}$ takes these enriched tokens and the residual tokens $\delta^{(t)}_{\texttt{emb}}$ in order to add the residual information as in Eq.~\ref{eq:codec-eq}.
These final features $\hat{X}^{(t)}_P$ should be similar to the image features extracted from the raw $\hat{I}^{(t)}$.
The process can be summarized as:
\begin{align}
X^{(t - 1)}_I = & \phi_{RGB}(I^{(t - 1)}) \\
\hat{X}^{(t - 1)}_{ \texttt{warped}} = &\theta_{\texttt{ref}}(X^{(t - 1)}_I, \tau^{(t)}_{\texttt{tok}})\\
\hat{X}^{(t)}_P =&\theta_{\texttt{warped}}(\hat{X}^{(t - 1)}_{\texttt{warped}}, \delta^{(t)}_{\texttt{tok}})
\enspace .
\end{align}

To align $\Delta$-tokens with image tokens, we apply patch-wise regression against the outputs of a frozen vision encoder. Let $X^{(t)}_I = \phi_{\text{RGB}}(\hat{I}^{(t)}) \in \mathbb{R}^{M \times d}$ denote the tokens of the ground-truth target frame. We minimize:
\begin{equation}
\mathcal{L}_{\text{MSE}} = \frac{1}{M} \sum_{i=1}^M \big\| X^{(t)}_I(i) - \hat{X}^{(t)}_P(i) \big\|_2^2.
\end{equation}

Unlike global contrastive losses, this fine-grained objective encourages spatially consistent alignment across patches. As a result, the $\Delta$-Encoder is able to produce representations that are more closely aligned with the RGB token space, aiming to improve the integration of I- and P-frames during downstream VideoLM training (see Supp.; skipping pre-training and single stage end-to-end training weakens performance).

\PAR{Integration into VideoLMs.} After pre-training, we integrate the $\Delta$-encoder $\phi_{\Delta}$ into the VideoLM pipeline for full fine-tuning and interleave the tokens coming from I/ P-frames.
Note that the Reference and Warped transformers from the $\Delta$-encoder pre-training stage are \textbf{\textit{not}} used at this stage, so no RGB reference frames are processed for the P-frames when training the language model.
This yields a substantial compute and memory reduction (shown in Sec.~\ref{sec:runtime_memory_eval}).
The LLM architecture and training objective remain unchanged (standard instruction tuning / next-token prediction loss). The $\Delta$-encoder adds fewer than $15$M parameters, representing negligible overhead relative to the $7$B LLM.

\section{Experiments}
\label{sec:experiment}

\subsection{Experimental Setup}
\label{sec:exp_setup}

\PAR{Training Pipeline.} For simplicity, we re-encode videos to the MPEG-4 codec at $30$ FPS with a GOP size of $240$ frames and use a fusion size of $s=30$, yielding an effective rate of $1$ FPS. We use LLaVA-Video-7B~\cite{zhang2024videoinstructiontuningsynthetic} as the base VideoLM, which consists of SigLIP~\cite{siglip} as the vision encoder and Qwen2-7B~\cite{yang2024qwen2technicalreport} as the language model. First, we pre-train the $\Delta$-encoder on frame pairs sampled from videos in the LLaVA-Video-178K~\cite{zhang2024videoinstructiontuningsynthetic}. Second, we fine-tune the full VideoLM on the same dataset, comprising a total of $1.39$M QA instruction tuning samples. We use a learning rate of $1\mathrm{e}{-5}$, an effective total batch size of $128$, and train for $21$K GPU hours ($64$ A100-80G for $14$ days).

\PAR{Evaluation Benchmarks.} We comprehensively evaluate our method across $14$ video benchmarks spanning four categories: (i)~\textit{general video QA}: PerceptionTest~\cite{patraucean2023perception}, NextQA~\cite{xiao2021next}, ActivityNet-QA~\cite{yu2019activityqa}, and VideoMME~\cite{fu2024videomme}; (ii)~\textit{temporal reasoning}: TempCompass~\cite{liu2024tempcompass}, TOMATO~\cite{shangguan2024tomatoassessingvisualtemporal}, CVRR-ES~\cite{Khattak2024cvrres} and MVBench~\cite{li2024mvbench}; (iii)~\textit{long-form and instruction-following}: LongVideoBench~\cite{wu2024longvideobench}, LVBench~\cite{wang2024lvbench}, Video-TT~\cite{zhang2025videothinkingtestholistic}, and Video-MMMU~\cite{hu2025videommmu}; and (iv)~\textit{spatial scene understanding}: ScanQA~\cite{azuma2022scanqa} and SQA3D~\cite{ma2023sqa3d}. Our primary comparison is with LLaVA-Video-7B~\cite{zhang2024videoinstructiontuningsynthetic}, which serves as our base model. We additionally compare with various similar open-source approaches. Following standard practice, we use \texttt{lmms-eval}~\cite{zhang2024lmms} for evaluation. ActNet-QA~\cite{yu2019activityqa} uses GPT-based evaluation; we report scores using \textit{gpt-4o-2024-11-20} in Tab.~\ref{tab:main_results} and Azure deployed version of \textit{gpt-3.5-turbo-0613} in Tab.~\ref{tab:video_qa_general} for fair comparison with prior work.

\PAR{Frame Sampling.} Most open-source VideoLMs with a 32K token window sample $64$ frames per video, regardless of length~\cite{liu2024oryx,zhang2024videoinstructiontuningsynthetic}. As a result, unlike proprietary models, they cannot process videos at $1$~FPS for videos longer than $64$s. Our framework decouples frame rate from token budget: we sample at $1$~FPS (counting both I- and P-frames), with the number of keyframes per GOP controlling the token cost rather than the temporal coverage. Concretely, each GOP contributes one I-frame ($M$ tokens) plus up to seven P-frames ($N=8$ tokens each), so the per-GOP cost ranges from $M + 7N$ (1 keyframe) to $4M + 4N$ (4 keyframes) depending on configuration. This allows the token budget to scale naturally with video duration. In our experiments, for fair comparison, we sample at $1$ FPS up to $64$ GOPs. For videos longer than $64$ GOPs ($= 512$s), we perform uniform GOP sampling and encode only I-/P-frames within each sampled GOP.

\subsection{Effectiveness of $\Delta$-tokens}
\label{sec:main_results}
We first evaluate whether the proposed $\Delta$-tokens derived from P-frames provide understanding capabilities comparable to dense RGB representations. 
Tab.~\ref{tab:main_results} summarizes results on three benchmarks under different sampling configurations. Across all settings, our codec-aware representation achieves consistent improvements in accuracy compared to the same-sampling baseline, and helps close the gap to the  denser sampling configuration while using significantly fewer tokens. Remarkably, even under aggressive compression (e.g., $1$ keyframe per GOP), our model maintains strong performance with over an order-of-magnitude token reduction. This demonstrates that the $\Delta$-encoder successfully captures motion and appearance cues critical for temporal reasoning (see Supp.; zeroing out $\Delta$-tokens at inference time confirms the VideoLM actively utilizes them and the same scaling trend holds when our model serves as the keyframe-only baseline). At higher frame densities (e.g., $4$ keyframes per GOP), our model not only matches but often surpasses the performance of the $64$-frame LLaVA-Video-7B~\cite{zhang2024videoinstructiontuningsynthetic} baseline, despite using only a fraction of its tokens. The largest relative gains are observed on PerceptionTest~\cite{patraucean2023perception} ($+6.7\%$) and NextQA~\cite{xiao2021next} ($+1.6\%$), indicating stronger reasoning under a limited token budget. These results confirm that our codec-aware tokenization preserves temporal semantics and fine-grained dynamics. As shown in the accompanying plots, our method consistently advances the Pareto-optimal frontier across all three benchmarks, achieving competitive or superior accuracy at substantially reduced token counts.

\begin{table*}[t]
\centering
\scriptsize
\caption{\textbf{Token Efficiency vs. Accuracy in Video QA.} We report the performance of LLaVA-Video (7B) at different number of keyframes per GOP, as well as in the default setup of selecting 64 keyframes regardless of video length. For each setting, we also report the performance of our method using the same keyframes as I-frames and the remaining frames in the GOP as P-frames.
We report accuracy (Acc., \%) and the \% of tokens used compared to the default setup (64 keyframes).
Our method only adds a low number of $\Delta$-tokens compared to its associated baseline and these help close the gap compared to the next baseline that is using a significantly larger number of tokens.}
\label{tab:main_results}
\setlength{\tabcolsep}{1pt}
\begin{minipage}{0.70\linewidth}
\resizebox{\linewidth}{!}{ 
\begin{tabular}{l c c l c l c l}
\toprule
 \multirow{2.5}{*}{\textbf{Model}} & \multirow{2.5}{*}{\textbf{Sampling}} &
\multicolumn{2}{c}{\textbf{PerceptionTest}} &
\multicolumn{2}{c}{\textbf{NextQA}} &
\multicolumn{2}{c}{\textbf{ActNet-QA}} \\
\cmidrule(lr){3-4} \cmidrule(lr){5-6} \cmidrule(lr){7-8}
 & & Token (\%) & Acc. (\%) & Token (\%) & Acc. (\%) & Token (\%) & Acc. (\%) \\
\midrule

LLaVA-Video & ~~1 keyframe / GOP
 & $\textcolor{white}{00}5.3$  & $60.4$ & $\textcolor{white}{00}8.3$ & $77.9$ & $\textcolor{white}{0}21.6$ & $61.7$ \\
 
\textcolor{copecolor}{\textbf{\textit{Ours}}} & \textcolor{blue}{+ 7 P-frames / GOP} 
 & $\textcolor{white}{00}6.9$ 
 & $65.5\smash{\raisebox{+1.9ex}{\scriptsize\textcolor{ForestGreen}{+5.1}}}$ 
 & $\textcolor{white}{0}11.2$ 
 & $78.3\smash{\raisebox{+1.9ex}{\scriptsize\textcolor{ForestGreen}{+0.4}}}$ 
 & $\textcolor{white}{0}29.7$ 
 & $62.3\smash{\raisebox{+1.9ex}{\scriptsize\textcolor{ForestGreen}{+0.6}}}$ \\
 
LLaVA-Video & ~2 keyframes / GOP
 & $\textcolor{white}{00}9.7$ & $62.1$ & $\textcolor{white}{0}15.9$ & $79.2$ & $\textcolor{white}{0}42.5$ & $62.9$ \\
 
\textcolor{copecolor}{\textbf{\textit{Ours}}} & \textcolor{blue}{+ 6 P-frames / GOP}
 & $\textcolor{white}{0}11.3$ 
 & $68.7\smash{\raisebox{+1.9ex}{\scriptsize\textcolor{ForestGreen}{+6.6}}}$ 
 & $\textcolor{white}{0}18.6$ 
 & $80.4\smash{\raisebox{+1.9ex}{\scriptsize\textcolor{ForestGreen}{+1.2}}}$ 
 & $\textcolor{white}{0}50.4$ 
 & $63.6\smash{\raisebox{+1.9ex}{\scriptsize\textcolor{ForestGreen}{+0.7}}}$ \\
 
LLaVA-Video & ~4 keyframes / GOP
 & $\textcolor{white}{0}18.6$ & $63.6$ & $\textcolor{white}{0}31.0$ & $80.5$ & $\textcolor{white}{0}84.3$ & $63.6$ \\
 
\textcolor{copecolor}{\textbf{\textit{Ours}}} & \textcolor{blue}{+ 4 P-frames / GOP}
 & $\textcolor{white}{0}20.1$ 
 & $\small\fs 70.3 \smash{\raisebox{+1.9ex}{\scriptsize\textcolor{ForestGreen}{+6.7}}}$ 
 & $\textcolor{white}{0}33.5$ 
 & $\nd 82.1 \smash{\raisebox{+1.9ex}{\scriptsize\textcolor{ForestGreen}{+1.6}}}$ 
 & $\textcolor{white}{0}91.7$ 
 & $\small\fs 64.8 \smash{\raisebox{+1.9ex}{\scriptsize\textcolor{ForestGreen}{+1.2}}}$ \\
 
\midrule
\midrule

LLaVA-Video & 64 keyframes total
 & $100$ & $\nd 67.9$ & $100$ & $\small \fs 83.2$ & $100$ & $\nd 64.1$ \\

\bottomrule
\end{tabular}}
\end{minipage}
~
\begin{minipage}{.27\linewidth}
\includegraphics[width=\linewidth]{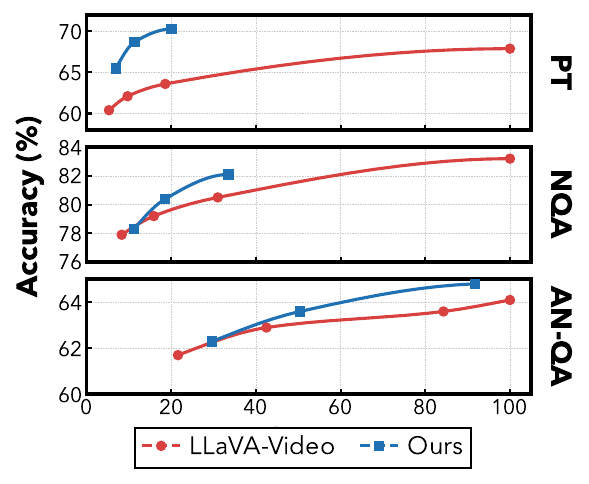}
\end{minipage}
% \vspace{-6pt}
\end{table*}

\subsection{Comparison with Current Approaches}
\label{sec:compare_curr}

We benchmark \project{} against a broad range of both open-source and proprietary VideoLMs. We consider four groups of tasks (see Supp. for a detailed description of the benchmarks): (i) general video QA (Tab.~\ref{tab:video_qa_general}), (ii) temporal and motion reasoning (Tab.~\ref{tab:video_qa_temporal_long} (a)), (iii) long-form and instruction following (Tab.~\ref{tab:video_qa_temporal_long} (b)), and (iv) spatial scene understanding (see Supp.; our method performs on par with 2D VLMs using only $25$\% of tokens).

\begin{table}[t]
    \centering
    \scriptsize
    \caption{\textbf{General video understanding benchmarks.} The best results among open-source methods are highlighted as \colorbox{colorFst}{\bf first}, \colorbox{colorSnd}{second}, and \colorbox{colorTrd}{third}. Our model achieves state-of-the-art performance among open-source video language models of comparable scale. Notably, it does so while operating on a fraction of the visual tokens, confirming that codec primitives preserve the semantic and temporal cues needed for diverse understanding tasks.}
    \label{tab:video_qa_general}
    \begin{tabular}{lccccc}
    \toprule
    \multirow{2}{*}{\textbf{Model}} & \textbf{PerceptionTest} & \textbf{NextQA} & \textbf{ActivityNet-QA} & \multicolumn{2}{c}{\textbf{VideoMME}} \\
    & val & mc & test & w/o sub & w sub \\
    \midrule
    \multicolumn{6}{l}{\cellcolor[HTML]{EEEEEE}{\textit{Proprietary Models}}} \\
    GPT-5~\cite{gpt5} & - & $86.3$ & -  & $83.3$ & $86.9$ \\
    Gemini 3 Pro~\cite{gemini3}
      & - & $84.3$ & - & $88.6$ & $87.5$ \\
    Gemini 2.5 Pro~\cite{comanici2025gemini}
      & - & $85.3$ & - & $87.8$ & $87.8$ \\
    Claude Sonnet 4.5~\cite{anthropic2025sonnet}
      & - & $79.2$ & - & $74.2$ & $80.5$ \\
    \midrule
    \multicolumn{6}{l}{\cellcolor[HTML]{EEEEEE}{\textit{Open-Source VideoLMs}}} \\
    Video-LaVIT~\cite{videolavit} & $47.9$ & - & $50.1$ & - & - \\
    EMA-7B~\cite{ema2025} & - & - & $52.1$ & $53.4$ & $58.4$ \\
    VILA-40B~\cite{lin2023vila}
    & $54.0$ & $67.9$ & $58.0$ & $60.1$ & $61.1$ \\
    LongVA-7B~\cite{zhang2024longva}
      & - & $68.3$ & $50.0$ & $52.6$ & $54.3$ \\
    IXC-2.5-7B~\cite{internlmxcomposer2_5}
      & $34.4$ & $71.0$ & $52.8$ & $55.8$ & $58.8$ \\
    LLaVA-OV-7B~\cite{li2024llava}
      & $57.1$ & $79.4$ & $\nd 56.6$ & $58.2$ & $61.5$ \\ 
    Apollo-7B~\cite{apollo} & $67.3$ & - & - & $\rd 61.3$ & $\rd 63.3$ \\
    Oryx-7B~\cite{liu2024oryx} & $\nd 68.6$ & $\rd 81.9$ & - & $58.3$ & $62.6$ \\
    LLaVA-Video-7B~\cite{zhang2024videoinstructiontuningsynthetic}
      & $\rd 67.9$ & $\fs 83.2$ & $\rd 56.5$ & $\fs 63.3$ & $\fs 69.7$ \\
    \arrayrulecolor{black!30}\midrule
    \textcolor{copecolor}{\textbf{\textit{Ours-7B}}} & $\fs 70.3$ & $\nd 82.1$ & $\fs 60.3$ & $\nd 61.9$ & $\nd 69.4$ \\
   \bottomrule
    \end{tabular}
    \vspace{-10pt}
\end{table}

\PAR{General Video QA.} As shown in Tab.~\ref{tab:video_qa_general}, despite being trained on a smaller corpus than most competing models, our codec-aware formulation achieves competitive or superior results across all major benchmarks. By leveraging motion vectors and residuals directly from the compressed stream, \project{} encodes substantially more frames within the same token budget, enhancing temporal coverage without sacrificing spatial fidelity. On PerceptionTest and ActivityNet-QA, our model yields the highest accuracy among all open-source models, indicating improved motion and appearance reasoning due to the $\Delta$-encoder's temporally grounded representation. The performance gap observed on select benchmarks such as VideoMME is attributable to our smaller training corpus rather than the codec formulation itself; we discuss this in detail in the Supp., where matched-data comparisons consistently favor our method. Notably, among codec-based approaches, \project{} substantially outperforms both Video-LaVIT~\cite{videolavit} and EMA~\cite{ema2025} across all reported benchmarks, validating that preserving both residuals and temporal ordering yields stronger video-language representations than fixed-length GOP summaries or motion-only tokenization.

\begin{table*}[t]
    \centering
    \scriptsize
    \caption{\small
    (a) \textbf{Temporal reasoning and motion understanding benchmarks.} \project{} achieves the highest accuracy on TempCompass, TOMATO, and CVRR-ES, confirming that codec primitives provide a strong inductive bias for temporal reasoning. (b) \textbf{Long-form and instruction-following benchmarks.} \project{} performs better than other open-source models on Video-TT, Video-MMMU, and LVBench, demonstrating compact $\Delta$-tokens effectively scale to longer temporal contexts and complex instruction-following.}
    \label{tab:video_qa_temporal_long}
    \resizebox{\textwidth}{!}{%
        \begin{tabular}[t]{c @{\hspace{3em}} c}
            \begin{tabular}[t]{lcccc}
            \toprule
            \multirow{2}{*}{\textbf{Model}} & \textbf{TempCompass} & \textbf{Tomato} & \textbf{CVRR-ES} & \textbf{MVBench} \\
            & test MCQ & test & test & test \\
            \midrule
            \multicolumn{5}{l}{\cellcolor[HTML]{EEEEEE}{\textit{Proprietary Models}}} \\
            GPT-5~\cite{gpt5}
              & $80.4$ & $53.0$ & - & $74.1$ \\
            Gemini 3 Pro~\cite{gemini3}
              & $82.8$ & $48.3$ & - & $70.4$ \\
            Gemini 2.5 Pro~\cite{comanici2025gemini}
              & $81.9$ & $48.6$ & - & $70.6$ \\
            Claude Sonnet 4.5~\cite{anthropic2025sonnet}
              & $72.8$ & $39.6$ & - & $62.1$ \\
            \midrule
            \multicolumn{5}{l}{\cellcolor[HTML]{EEEEEE}{\textit{Open-Source VideoLMs}}} \\
            LongVA-7B~\cite{zhang2024longva} & $56.9$ & - & - & - \\
            VideoLLaMA2-7B~\cite{damonlpsg2024videollama2} & - & $18.5$ & $21.6$ & $54.6$ \\
            InternVL2-8B~\cite{chen2024internvl2.5} & $65.3$ & $21.7$ & - & $\nd 65.8$ \\
            VideoChat2-7B~\cite{li2024mvbench} & $45.5$ & - & - & $51.1$ \\
            VideoCCAM-9B~\cite{fei2024videoccamenhancingvideolanguageunderstanding} 
            & - & $\rd 27.0$ & - & $\rd 64.6$ \\
            IXC-2.5-7B~\cite{internlmxcomposer2_5}
            & $\nd 67.1$ & - & - & $\fs 69.1$ \\
            LLaVA-OV-7B~\cite{li2024llava}
            & $64.8$ & $\rd 25.5$ & $\rd 42.6$ & $56.7$ \\ 
            Apollo-7B~\cite{apollo} & $64.9$ & - & - & - \\
            LLaVA-Video-7B~\cite{zhang2024videoinstructiontuningsynthetic} 
            & $\rd 66.6$ & $24.9$ & $\nd 43.6$ & $58.6$ \\
            \arrayrulecolor{black!30}\midrule
            \textcolor{copecolor}{\textbf{\textit{Ours-7B}}}
            & $\fs 68.9$ & $\fs 28.3$ & $\fs 49.4$ & $61.9$ \\
            \bottomrule
            \end{tabular}
            & 
            \begin{tabular}[t]{lcccc}
            \toprule
            \multirow{2}{*}{\textbf{Model}} & \textbf{Video-TT} & \textbf{Video-MMMU} & \textbf{LVBench} & \textbf{LongVideoBench} \\
            & mc & test & test & val \\
            \midrule
            \multicolumn{5}{l}{\cellcolor[HTML]{EEEEEE}{\textit{Proprietary Models}}} \\
            GPT-5~\cite{gpt5}
              & - & - & $68.8$ & $72.6$ \\
            Gemini-3-Pro~\cite{gemini3}
              & - & - & $78.0$ & $75.9$ \\
            Gemini-2.5-Pro~\cite{comanici2025gemini}
              & - & - & $78.4$ & $76.8$ \\
            Claude-Sonnet-4.5~\cite{anthropic2025sonnet}
              & - & - & $50.5$ & $65.1$ \\
            \midrule
            \multicolumn{5}{l}{\cellcolor[HTML]{EEEEEE}{\textit{Open-Source VideoLMs}}} \\
            EMA-7B~\cite{ema2025} & - & - & - & $47.0$ \\
            LongVA-7B~\cite{zhang2024longva} & - & $23.9$ & - & - \\
            Kangaroo-8B~\cite{kangaroogroup} & - & - & $39.4$ & $54.8$ \\
            mPLUG-Owl3-7B~\cite{ye2024mplugowl3longimagesequenceunderstanding} & - & - & $\rd 43.5$ & $52.1$ \\
            PLLaVA-34B~\cite{xu2024pllava} & - & - & $26.1$ & $53.2$ \\
            InternVL2-8B~\cite{chen2024internvl2.5} & - & $\nd 37.4$ & - & $54.6$ \\
            TimeMarker~\cite{timemarker} & - & - & $41.3$ & $56.3$ \\
            LLaVA-OV-7B~\cite{li2024llava} 
            & $\nd 44.0$ & $33.9$ & $38.1$ & $\rd 56.5$ \\
            LLaVA-Video-7B~\cite{zhang2024videoinstructiontuningsynthetic} 
            & $\rd 41.8$ & $\rd 36.1$ & $\nd 44.2$ & $\fs 58.2$ \\
            \arrayrulecolor{black!30}\midrule
            \textcolor{copecolor}{\textbf{\textit{Ours-7B}}}
            & $\fs 45.5$ & $\fs 38.2$ & $\fs 46.4$ & $\nd 56.9$ \\
            \bottomrule
            \end{tabular} \\
            \rule{0pt}{5ex}
            \small (a) & \small (b)
        \end{tabular}
    }
\end{table*}

\PAR{Temporal And Motion Reasoning.} Tab.~\ref{tab:video_qa_temporal_long} (a) evaluates benchmarks specifically designed to probe temporal understanding. On TempCompass~\cite{liu2024tempcompass},
TOMATO~\cite{shangguan2024tomatoassessingvisualtemporal} and CVRR-ES~\cite{Khattak2024cvrres}, \project{} achieves the highest accuracy among all open-source models. This confirms that the explicit encoding of motion vectors and residuals provides a stronger temporal signal than dense RGB frame processing. These gains are notable given that temporal reasoning requires precisely the kind of fine-grained inter-frame dynamics that codec primitives natively capture. On MVBench, which emphasizes pictorial semantic understanding over temporal dynamics, our model improves over LLaVA-Video-7B by $3.0\%$. The remaining gap to the top-ranked models  reflects the base model's capacity and training data mixture rather than a codec-specific limitation.

\PAR{Long-form and Instruction-following.} Tab.~\ref{tab:video_qa_temporal_long} (b) reports results on benchmarks requiring understanding of extended video sequences and complex instructions. \project{} achieves the best results among open-source models on Video-TT~\cite{zhang2025videothinkingtestholistic}, Video-MMMU~\cite{hu2025videommmu}, and LVBench~\cite{wang2024lvbench}, while remaining competitive on LongVideoBench~\cite{wu2024longvideobench}. The improvement is especially pronounced against EMA~\cite{ema2025}, even though we train on less data ($1.39$M video-only vs. $2.4$M video+image QA samples) and do not need to increase the GOP count at training time to handle longer videos. These results suggest that preserving residuals and explicit temporal ordering plays an important role for long-form understanding. Overall, compressing P-frames into compact $\Delta$-tokens enables substantially more temporal coverage within the same token budget, supporting both fine-grained temporal reasoning and efficient use of the context window.

\subsection{Runtime and Memory}
\label{sec:runtime_memory_eval}
\begin{figure*}[t]
    \begin{minipage}[t]{0.48\linewidth}
        \vspace{0pt}
        \centering
        \small
        \resizebox{\linewidth}{!}{
        \setlength{\tabcolsep}{3pt}
        \begin{tabular}{lcc}
        \toprule
        \textbf{Model / Sampling} & \textbf{TTFT (s)} & \textbf{E2EL (s)} \\
        \midrule
        LLaVA-Video-7B ($64$ kf) & $2.39$ & $3.78$ \\
        \midrule
        \midrule
        \multicolumn{3}{l}{\textcolor{copecolor}{\textbf{\textit{Ours-7B}}}} \\
        \makecell{$1$ keyframe per GOP\\($8$ kf + $56$ P-frames)} & $\textbf{0.33}$ & $\textbf{1.66}$ \\ \hdashline
        \makecell{$2$ keyframes per GOP\\($16$ kf + $48$ P-frames)} & $0.51$ & $1.71$ \\ \hdashline
        \makecell{$4$ keyframes per GOP\\($32$ kf + $32$ P-frames)} & $0.90$ & $2.28$ \\
        \arrayrulecolor{black}\bottomrule
        \end{tabular}
        }
    \end{minipage}
    \hfill
    \begin{minipage}[t]{0.48\linewidth}
        \vspace{0pt}
        \centering
        \includegraphics[width=\linewidth]{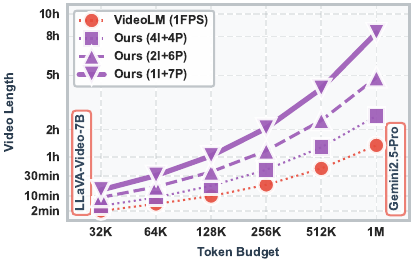}
    \end{minipage}
    \vspace{4pt}
    \makebox[0.48\textwidth]{\small (a)}\hfill\makebox[0.48\textwidth]{\small (b)}
    \caption{\small
    (a) \textbf{Runtime comparison:} TTFT and E2EL for generating $64$ text tokens at $1$ FPS on a single GPU. (b) \textbf{Video length vs.\ token budget:} Token budget is shown on a log scale; dashed lines mark evaluated budgets. $\Delta$-token representation enables scaling to significantly longer videos without exceeding context limits.}
    \label{fig:runtime_and_memory}
    % \vspace{-10pt}
\end{figure*}

Beyond competitive performance in accuracy, \project{} provides critical efficiency gains during inference. Fig.~\ref{fig:runtime_and_memory} (a) reports \textit{time-to-first-token} (TTFT) and \textit{end-to-end-latency} (E2EL) to generate $64$ text tokens, measured on a single consumer-grade GPU at $1$ FPS video input. Compared to the 64-frame LLaVA-Video-7B baseline, our most compact configuration (1 keyframe per GOP) achieves a $86.2\%$ reduction in TTFT and a $56.1\%$ faster E2EL. This improvement stems from the reduced visual embedding load (only I-frames require full RGB encoding) and the shorter overall sequence length processed by the LLM due to the $\Delta$-tokens. We highlight the scalability of the computational advantage of the $\Delta$-token formulation in Fig.~\ref{fig:runtime_and_memory} (b). Standard dense RGB sampling saturates quickly, limiting coverage to short sequences as memory constraints are rapidly encountered. In contrast, \project{} exhibits a highly efficient relationship between video length and token budget. The token efficiency enables scaling to sequences previously inaccessible to open-source models; our most compact configuration allows for the processing of videos up to $8$ hours in duration (at $1$ FPS) within a $1$M token context, demonstrating an order-of-magnitude increase in processing capability over the baseline. The benefits become increasingly significant for longer sequences, as the quadratic attention cost amplifies reductions in token count. Moreover, the architecture uses standard transformer components with well-established scaling properties. Together, these results confirm that codec-aware tokenization is not just semantically sound but is a necessity for enabling fast inference and comprehensive long-form video coverage without needing architectural modifications or additional hardware.

\section{Conclusion}
\label{sec:conclusion}

\PAR{Limitations and Future Work.} Our current approach focuses on I- and P-frames, lacking support for B-frames and their complex non-causal dependencies. One option to address this would be using the decode order instead of the render order. Furthermore, we currently operate on a tensorized version of the codec primitives. For future work, it would be interesting to stay closer to the raw codec primitives, by operating directly on sets of block-wise motion vectors and quantized DCT coefficients, which could offer even better computational and token efficiency. Finally, we use a fixed P-frame fusion window, which is suboptimal for tasks with varying motion. Exploring sensitivity to codec type, bitrate, and encoding quality remains an open question.

\PAR{Conclusion.} Through comprehensive evaluation across 14 video understanding benchmarks, we have demonstrated that codec-aware tokenization offers a compelling alternative to traditional keyframe sampling and outperforms prior codec-based approaches for VideoLMs. By leveraging the information natively encoded by video compression algorithms (i.e., motion vectors and residuals), we achieve substantial efficiency gains while maintaining competitive performance. Notably, our approach reduces time-to-first-token by a significant margin (up to $86\%$) which is essential for real-time applications. As models scale towards larger context windows, our approach becomes increasingly valuable, enabling richer temporal representations with much lower computational overhead than traditional sparse keyframe sampling. This work positions codec-based methods as a practical and efficient foundation for scaling future VideoLMs.
\section{Acknowledgements}

The authors would like to thank (in alphabetical order): Isar Meijer and  Krzysztof Waraksa from Microsoft for help with training pipeline setup; Kevin Qu, Tao Sun and Jianhao Zheng from Stanford for feedback at different stages of the project.
\setcounter{section}{0}
\counterwithin{table}{section}
\renewcommand{\thesection}{\Alph{section}}

\clearpage
\appendix
\section*{Appendix}

\noindent\textit{In the appendix, we provide the following:}

{\itshape
\begin{enumerate}
     \item Video decoding illustration (Sec.~\ref{sec:supp_codec_details})
     \item Details about $\Delta$-encoder (Sec.~\ref{sec:suppl_delta_enc_details})
     \item Additional training details (Sec.~\ref{sec:suppl_training_details})
     \item Details on training data and evaluation benchmarks (Sec.~\ref{sec:training_data_eval_benchmark_detail})
    \item Scale of training data (Sec.~\ref{sec:training_data_scale})
    \item Spatial video question-answering (Sec.~\ref{sec:spatial_qa})
    \item Comparison with token pruning (Sec.~\ref{sec:token_compression})
    \item Ablation study (Sec.~\ref{sec:ablation_study})
    \begin{itemize}
        \item Varying the number of $\Delta$-tokens (Sec.~\ref{subsec:var-delta})
        \item Two-Stage training (Sec.~\ref{subsec:two-stage})
        \item Are $\Delta$-tokens used by the VideoLM? (Sec.~\ref{subsec:delta-tok-used})
        \item Benefits of codec primitives (Sec.~\ref{subsec:controlled-ablation})
        \item Scaling to higher frame rates (Sec.~\ref{subsec:higher-fps})
        \item Next-frame retrieval using the $\Delta$-encoder (Sec.~\ref{subsec:next_frame_retrieve})
    \end{itemize}
    \item Qualitative results (Sec.~\ref{sec:quals_results})
\end{enumerate}
}

\section{Video Decoding Illustration}
\label{sec:supp_codec_details}
\begin{figure}[h]
    \centering
    \includegraphics[width=\linewidth]{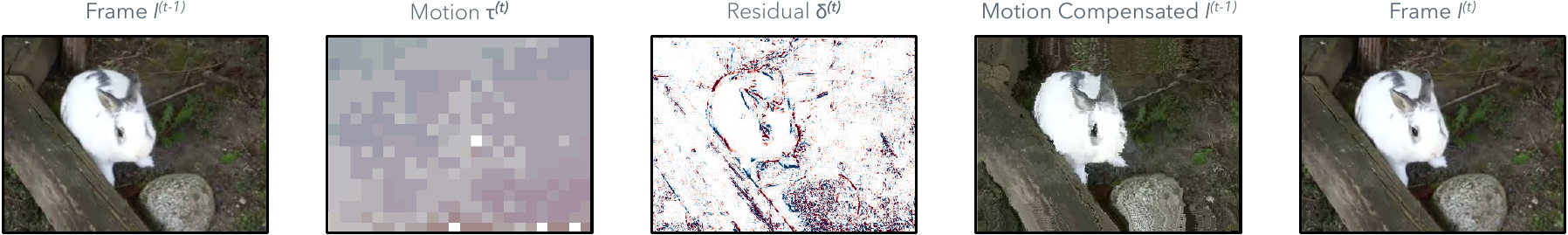}
    \caption{
    \textbf{Codec primer.}
    We visualize from left to right: the previous frame, the motion vectors and residuals between previous and current frame, the intermediate reconstruction after motion compensation, and the final result after adding the residuals.
    }
\label{fig:codec_primer}
\end{figure}

As mentioned in Sec.~\ref{sec:preliminaries}, the size of each Group of Pictures (GOP) is usually decided adaptively by the codec and video encoder depending on the motion or change. At encoding time, the developer can provide an upper bound for the GOP size, or even fix it.
For example, if the GOP size is fixed at $240$, the first frame of each GOP is an I-frame and the rest $239$ frames are P-frames, each consisting of motion vectors and residuals. During video decoding, the I-frame or reconstructed frame at the previous timestamp $t - 1$ is first moved using the motion vectors and then the residuals are added to get the RGB frame at the current timestep $t$, as described in Eq.~\ref{eq:codec-eq}. We visualize an example of the decoding process in Fig.~\ref{fig:codec_primer}.

\section{Details about $\Delta$-Encoder}
\label{sec:suppl_delta_enc_details}
The $\Delta$-Encoder converts codec primitives, motion vectors and residuals, into a compact set of $N = K_\tau + K_\delta$ tokens aligned with the vision encoder's embedding space. The entire module is lightweight ($<15$M parameters) and operates purely in the compressed domain during VideoLM fine-tuning. Here, we provide additional architectural and implementation details. 

\PAR{Motion Vector Branch.} Given motion vectors $\tau^{(t)} \in \mathbb{Z}^{H \times W \times 2}$, we first perform min–max normalization to map all values to $[-1,1]$.
We then ``patchify'' the motion-field into non-overlapping $16 \times 16$ blocks, yielding a grid of size $H_G \times W_G$ where $H_G=\nicefrac{H}{16}$ and $W_G=\nicefrac{W}{16}$.
Each patch is flattened into a vector of dimension $16^2\!\times\!2$, and a lightweight two-layer MLP with shared weights is applied independently to each patch, producing per-block embeddings with the same feature dimension as the vision encoder ($d=1152$).
To aggregate these features, we employ a transformer equipped with $K_\tau$ learnable query tokens.
These tokens are concatenated with the motion features along the sequence length dimension at the input of the transformer and then processed together using regular multi-headed attention layers.
The transformer contains $4$ layers with hidden dimension $d$, uses $9$ attention heads, and adopts PreNorm residual blocks throughout.

\PAR{Residual Branch.} Residuals $\delta^{(t)} \in \mathbb{R}^{H \times W \times C}$ are processed 
through a truncated ResNet-18 backbone (all convolutional layers up to the final global pooling)~\cite{He2015DeepRL}. The resulting spatial features share the same grid resolution as above $(H_G \times W_G)$. A second transformer, architecturally identical to the motion branch but with its own learned $K_\delta$ queries, is used to compress these features.

\PAR{Pre-training the $\Delta$-Encoder.} As mentioned in Sec.~\ref{sec:training_paradigm}, the $\Delta$-Encoder is augmented with two auxiliary transformer modules, \emph{reference} and \emph{warped} branches, that enable it to reconstruct the token representation of the target frame without decoding RGB pixels. These transformers are architecturally identical to the ones in the motion vector and residual branch.

\section{Additional Training Details}
\label{sec:suppl_training_details}
\PAR{Pre-training.} We pre-train the $\Delta$-encoder with the \textit{warped} and \textit{reference} branches together for two days using 16$\times$A100 GPUs, running 113K iterations with a global batch size of 1024 and a per-GPU learning rate of $6.25\times10^{-5}$, optimized with AdamW and a cosine scheduler with warmup steps $= 1000$.

\PAR{VLM Training.} To better adhere to the original RGB aligned latent space, we train the VideoLM with 4 keyframes + 4 P-frames per GOP, for $10.9$K steps on 64$\times$A100 GPUs with a global batch size of $128$. We keep the same hyperparameter settings as LLaVA-Video~\cite{zhang2024videoinstructiontuningsynthetic}. 

\section{Training Data and Evaluation Benchmarks}
\label{sec:training_data_eval_benchmark_detail}
\PAR{Training Data.} We train on
LLaVA-Video-178K~\cite{zhang2024videoinstructiontuningsynthetic}, a video
instruction-tuning dataset organized by duration ($0$--$30$s, $30$--$60$s, $1$--$2$min, $2$--$3$min) with sources from academic benchmarks and YouTube. It covers captioning, open-ended QA, and multiple-choice QA. ActivityNet-QA~\cite{yu2019activityqa}, NextQA~\cite{xiao2021next}, and
PerceptionTest~\cite{patraucean2023perception} are included as QA subsets; following prior work, we treat these together with the rest of LLaVA-Video-178K as a single training corpus totalling $1.39$M samples.

\PAR{General Video QA.} PerceptionTest~\cite{patraucean2023perception} tests fine-grained perceptual reasoning (memory, abstraction, physics) through multiple-choice questions
over short diagnostic videos. NextQA~\cite{xiao2021next} targets causal, temporal, and descriptive
reasoning on short clips via multiple-choice QA.
ActivityNet-QA~\cite{yu2019activityqa} poses open-ended questions about complex web videos spanning diverse activities, scored by GPT. VideoMME~\cite{fu2024videomme} covers short to long videos with questions on spatial, temporal, and semantic understanding, reported with and without subtitles.

\PAR{Temporal and Motion Reasoning.}
TempCompass~\cite{liu2024tempcompass} probes temporal understanding through questions about speed, direction, order, and attribute change. TOMATO~\cite{shangguan2024tomatoassessingvisualtemporal} tests action ordering, duration, and state transition reasoning via multiple-choice QA. CVRR-ES~\cite{Khattak2024cvrres} uses compositionally varied questions to expose systematic failure modes in video reasoning. MVBench~\cite{li2024mvbench} spans 20 temporal tasks including action sequence, scene transition, and object interaction recognition.

\PAR{Long-form and Instruction Following.}
LongVideoBench~\cite{wu2024longvideobench} tests understanding of long, interleaved video-language sequences with multi-step reasoning. LVBench~\cite{wang2024lvbench} targets long-form video comprehension with
videos exceeding several minutes. Video-TT~\cite{zhang2025videothinkingtestholistic} tests complex multi-step video reasoning and holistic understanding.
Video-MMMU~\cite{hu2025videommmu} tests knowledge acquisition from multi-discipline professional videos.

\section{Scale of Training Data}
\label{sec:training_data_scale}

As mentioned in Sec.~\ref{sec:compare_curr}, \project{} underperforms LLaVA-Video on select benchmarks, e.g., VideoMME~\cite{fu2024videomme}. This gap reflects training data scale and composition rather than the codec formulation. Tab.~\ref{tab:data_scale_comparison} isolates this effect by reporting results from the LLaVA-Video  paper~\cite{zhang2024videoinstructiontuningsynthetic}
across its incremental training stages. LLaVA-Video begins with $0.25$M samples (LLaVA-Hound) and progressively adds LLaVA-Video-178K ($1.58$M), three QA datasets ($1.64$M), and finally $1.1$M LLaVA-OneVision images ($2.74$M). Two observations stand out. First, VideoMME improves substantially with the image alignment data ($61.9 \to 63.4$), which our pipeline lacks
entirely. Second, adding the QA datasets reduces VideoMME ($63.2 \to 61.9$). Since this configuration closely mirrors our data mixture, and we observe similar VideoMME accuracy,  this confirms a data-mixture effect rather than a codec-specific limitation. We additionally include a
``sampled'' variant as reported in the LLaVA-Video
paper~\cite{zhang2024videoinstructiontuningsynthetic}, where LLaVA-Video is trained on a randomly sampled $1.08$M subset of LLaVA-Video-178K; our method
outperforms it by a wide margin across all benchmarks.

\begin{table}[h]
\centering 
\scriptsize
\caption{\textbf{Effect of training data scale.} LLaVA-Video rows are reported in~\cite{zhang2024videoinstructiontuningsynthetic} across incremental training stages. Against the most comparable configurations (+ 3 QA datasets and ``sampled''), our method outperforms or remains on par across benchmarks despite using fewer samples.}
\label{tab:data_scale_comparison} 
\begin{tabular}{lcccc} 
    \toprule 
    \textbf{Training Data} & \textbf{Total Data} & \textbf{NextQA} & \textbf{PerceptionTest} & \textbf{VideoMME} \\ 
    \midrule 
    \multicolumn{5}{l}{\cellcolor[HTML]{EEEEEE}{\textit{LLaVA-Video~\cite{zhang2024videoinstructiontuningsynthetic}}}} \\ 

    LLaVA-Hound & $0.25$M
    & $64.4$ & $51.4$ & $54.1$ \\ 

    \hspace{1.2em}+ LLaVA-Video-178K 
    & $1.58$M & $\rd 80.1$ & $57.1$ & $\nd 63.2$ \\ 

    \hspace{1.2em}+ 3 QA datasets 
    & $1.64$M & $\rd 80.1$ & $\nd 69.0$ & $\rd 61.9$ \\  

    \hspace{1.2em}+ LLaVA-OV (Images)
    & $2.74$M & $\fs 83.2$ & $\rd 67.9$ & $\fs 63.4$ \\  
    
    LLaVA-Video-178K (sampled) & $1.08$M
    & $73.2$ & $55.9$ & $59.6$ \\

    \arrayrulecolor{black!30}\midrule 
    \multicolumn{5}{l}{\cellcolor[HTML]{EEEEEE}{ \textcolor{copecolor}{\textit{Ours}}}} \\
    \makecell[l]{LLaVA-Video-178K + 3 QA datasets} & $1.39$M
    & $\nd 82.1$ & $\fs 70.3$ & $ \rd 61.9$ \\ 

    \arrayrulecolor{black}
    \bottomrule 
\end{tabular}
\end{table}

\section{Spatial Video Question Answering}
\label{sec:spatial_qa}

\begin{table*}[t]
    \centering\scriptsize
    \caption{\textbf{Evaluation of 3D question-answering} on SQA3D~\cite{ma2023sqa3d} and ScanQA~\cite{azuma2022scanqa}. ``Expert models'' are customized for specific tasks with task-oriented decoders. ``EM'' stands for top-1 exact match and ``EM-R'' means the refined exact match following~\cite{huang2023leo}. ``--'' indicates the number is not available. We show results for our method in zero-shot setup where it performs comparable to state-of-the-art VideoLMs despite using only $25\%$ of tokens. We further show results after fine-tuning where the performance increases significantly, even outperforming a significant number of 3D VLMs which employ additional inputs (e.g., point-clouds, camera poses).} 
    \label{tab:spatialqa}
    \setlength{\tabcolsep}{6.57pt}
    \resizebox{\linewidth}{!}{
    \begin{tabular}{lcc cc ccccc}
    \toprule
    \multirow{2.5}{*}{Method} & \multirow{2.5}{*}{\begin{tabular}[c]{@{}c@{}}Point\\ Encoder\end{tabular}} & \multirow{2.5}{*}{\begin{tabular}[c]{@{}c@{}}Vision\\ Encoder\end{tabular}} & \multicolumn{2}{c}{SQA3D$_{\text{test}}$} & \multicolumn{5}{c}{ScanQA$_{\text{val}}$} \\
    \cmidrule(lr){4-5}
    \cmidrule(lr){6-10}
    & & & EM & EM-R & CIDEr & BLEU-4 & METEOR & ROUGE & EM \\
    \midrule
    \multicolumn{10}{l}{\cellcolor[HTML]{EEEEEE}{ \textit{Expert Models}}} \\
    SQA3D~\cite{ma2023sqa3d} & \checkmark & \xmark & $46.6$ & -- & -- & -- & -- & -- & -- \\
    ScanQA~\cite{azuma2022scanqa} & \checkmark & \xmark & -- & -- & $64.9$ & $10.1$ & $13.1$ & $33.3$ & $21.1$ \\
    3D-VLP~\cite{jin2023context} & \checkmark & \xmark & -- & -- & -- & $11.2$ & $13.5$ & $34.5$ & $21.7$ \\
    3D-VisTA~\cite{zhu20233dvista} & \checkmark & \xmark & -- & -- & -- & -- & $13.9$ & $35.7$ & $22.4$ \\
    \midrule
    \multicolumn{10}{l}{\cellcolor[HTML]{EEEEEE}{ \textit{3D VLMs}}} \\
    Chat-3D~\cite{wang2023chat} & \checkmark & \xmark & -- & -- & $53.2$ & $6.4$ & $11.9$ & $28.5$ & -- \\
    3D-LLM~\cite{hong20233d} & \checkmark & \checkmark & -- & -- & $69.4$ & $12.0$ & $14.5$ & $35.7$ & $20.5$ \\
    Scene-LLM~\cite{fu2024scene} & \checkmark & \checkmark & $53.6$ & -- & $80.0$ & $11.7$ & $15.8$ & $35.9$ & $27.2$ \\
    LL3DA~\cite{chen2024ll3da} & \checkmark & \xmark & -- & -- & $76.8$ & -- & $15.9$ & $37.3$ & -- \\
    LEO~\cite{huang2023leo} & \checkmark & \checkmark & $50.0$ & $52.4$ & $80.0$ & $11.5$ & $16.2$ & $39.3$ & $21.5$ \\
    ChatScene~\cite{zhang2024chatscene} & \checkmark & \checkmark & $54.6$ & $57.5$ & $87.7$ & $14.3$ & $18.0$ & $41.6$ & $21.6$ \\
    Grounded 3D-LLM~\cite{chen2024grounded} & \checkmark & \checkmark & -- & -- & $72.7$ & $13.4$ & -- & -- & -- \\
    LLaVA-3D~\cite{zhu2024llava3d} & \xmark & \checkmark & $55.6$ & $\rd{57.6}$ & $91.7$ & $14.5$ & $\nd 20.7$ & $\nd 50.1$ & $27.0$ \\
    Video-3D-LLM~\cite{zheng2024video3dllm} & \xmark & \checkmark & $\nd 58.6$ & -- & $\nd 102.1$ & $\nd 16.4$ & $\rd{20.0}$ & $\rd{49.3}$ & $\nd 30.1$ \\
    Ross3D~\cite{wang2025ross3d} & \xmark & \checkmark & $\fs{63.0}$ & $\fs {65.7}$ & $\fs {107.0}$ & $\fs{17.9}$ & $\fs{20.9}$ & $\fs{50.7}$ & $\bf \fs{30.8}$ \\
    \midrule
    \multicolumn{10}{l}{\cellcolor[HTML]{EEEEEE}{ \textit{VideoLMs (Zero-shot)}}} \\
    InternVL2-8B~\cite{chen2024internvl2.5} & \xmark & \checkmark & $33.0$ & $45.3$ & $62.5$ & $3.3$ & $14.5$ & $34.3$ & -- \\
    Qwen2-VL-7B~\cite{Qwen2-VL} & \xmark & \checkmark & $40.7$ & $46.7$ & $53.9$ & $3.0$ & $11.4$ & $29.3$ & -- \\
    LLaVA-Video-7B~\cite{zhang2024videoinstructiontuningsynthetic} \textcolor{blue}{$100$\% tokens} & \xmark & \checkmark & $48.5$ & -- & $88.7$ & $3.1$ & $17.7$ & $44.6$ & -- \\
    \midrule
    \midrule
    \textcolor{copecolor}{\textit{\textbf{Ours-7B}} \textit{(Zero-shot)}} \textcolor{blue}{$25.78$\% tokens} & \xmark & \checkmark & $46.5$ & $49.8$ & $70.9$ & $7.1$ & $14.7$ & $38.2$ & -- \\
    \textcolor{copecolor}{\textit{\textbf{Ours-7B}} \textit{(Fine-tuned)}} \textcolor{blue}{$25.78$\% tokens} & \xmark & \checkmark & $\rd{56.6}$ & $\nd 59.3$ & $\rd{96.9}$ & $\rd 14.9$ & $19.1$ & $46.4$ & $\rd 27.5$ \\
    \bottomrule
    \end{tabular}}
\end{table*}

We evaluate our method on two standard 3D QA benchmarks:  (1) SQA3D~\cite{ma2023sqa3d} for situated reasoning, and  (2) ScanQA~\cite{azuma2022scanqa}, for spatial understanding. Both datasets require associating multi-view observations with 3D spatial structure, making them a natural testbed for assessing whether compressed-domain video cues can support geometry-aware reasoning.

\PAR{Training.} We follow the same training pipeline used in our video-language experiments.
However, since these videos are comparatively shorter in duration ($\sim15$s total), we re-encode them with a GOP size of $120$ at $30$ FPS with accumulation size $s=10$.
This leads to around 3-4 GOPs per video.
Following standard practice in 3D LMMs, we fine-tune our base model on ScanQA and SQA3D training split, without any dataset-specific heuristics or architectural modifications.
We train and evaluate with $6$ keyframes + $6$ P-frames per GOP to align better with $32$-RGB frame setting of the other models. 

\PAR{Results.} We report the results in Tab.~\ref{tab:spatialqa}. To maintain fairness with VideoLMs, we show results for both our original and fine-tuned model. Despite using $1/4$th of the number of tokens compared to LLaVA-Video-7B~\cite{zhang2024videoinstructiontuningsynthetic}, the original \project{} matches state-of-the-art VideoLMs. With fine-tuning, our performance becomes comparable to the leading 3D VLMs, despite our method not having access to camera poses or 3D point-clouds. However, we note that fine-tuning would likely benefit other VideoLMs similarly; the key takeaway is that codec-aware tokens retain sufficient spatial information for 3D reasoning at a fraction of the token cost.

\section{Comparison with Token Pruning}
\label{sec:token_compression}
As discussed in Sec.~\ref{sec:related_work}, token compression methods operate on dense vision tokens \textit{after} full RGB encoding. Although they reduce the LLM's token count, these still pay the full image embedding cost for every frame, which dominates prefill latency. In contrast, \project{} avoids full image encoding for most frames entirely by leveraging codec-level sparsity. Tab.~\ref{tab:token_compression_comparison} compares \project{} against FastV~\cite{fastv}, DyCoke~\cite{dycoke}, PLLaVA~\cite{xu2024pllava}, VisionZip~\cite{visionzip}, and LLaVA-Scissor~\cite{llavascissor}. All baselines use $50\%$
token budget of the default frame configuration, while our method enables up to 93\% lower token usage, as shown in Sec.~\ref{sec:exp_setup}.

\begin{table}[t]
	\centering
    \scriptsize
    \caption{\textbf{Comparison with token compression methods.} \project{} outperforms post-hoc pruning approaches across all three benchmarks while being faster in TTFT, as P-frames bypass the vision encoder entirely.}
    \label{tab:token_compression_comparison}
       \begin{tabular}{lccc}
            \toprule
            \textbf{Method} & \textbf{ActNet-QA} & \textbf{Next-QA} & \textbf{VideoMME} w/o sub \\
            \midrule
            FastV~\cite{fastv}  & $\nd 47.95$ & $\rd 81.1$ & $\nd 57.5$ \\
            DyCoke~\cite{dycoke} & $47.88$ & $\rd 81.1$ & $\rd 57.4$ \\
            PLLaVA~\cite{xu2024pllava} & $47.59$ & $81.0$ & $56.9$ \\
            VisionZip~\cite{visionzip} & $45.42$ & $78.5$ & $54.2$ \\
            LLaVA-Scissor~\cite{llavascissor} & $\rd 47.89$ & $\nd 81.2$ & $\rd 57.4$ \\
            Ours & $\fs 60.28$ & $\fs 82.1$ & $\fs 61.9$ \\
            \bottomrule
        \end{tabular}
\end{table}

\section{Ablation Study}
\label{sec:ablation_study}
To reduce carbon footprint and computation cost compared to the main high compute run, for the ablation experiments, we train our \project{} on the three QA datasets: PerceptionTest~\cite{patraucean2023perception}, NextQA~\cite{xiao2021next}, ActivityNet-QA~\cite{yu2019activityqa}, comprising $60$K samples for $2$ days on $16\times$A100.
Due to this, the results shown in Sec.~\ref{subsec:var-delta}, Sec.~\ref{subsec:two-stage}, and Sec.~\ref{subsec:delta-tok-used} are not directly comparable to these reported in Sec. $4$. We perform evaluation on the \textit{val} and \textit{test} splits of PerceptionTest~\cite{patraucean2023perception} and NextQA~\cite{xiao2021next} respectively.

\subsection{Varying the Number of $\Delta$-Tokens}
\label{subsec:var-delta}
The number of $\Delta$-tokens emitted per P-frame controls the expressive capacity of the codec branch. 
Fewer tokens encourage a more aggressively compressed representation, whereas larger token budgets allow the $\Delta$-Encoder to retain finer motion and appearance cues from the codec primitives. 
Tab.~\ref{tab:delta_token_comparison} reports the effect of varying this value while keeping all other model and sampling configurations fixed. We observe a consistent trend across both benchmarks. Moving from $2$ to $4$ $\Delta$-tokens yields a noticeable improvement, as the model benefits from an expanded latent space that can more faithfully capture motion and residual structure. Increasing to $8$ $\Delta$-tokens further improves performance, reaching the best overall results. Beyond this point, however, allocating more capacity produces diminishing returns: both NextQA and PerceptionTest remain nearly unchanged when increasing to $16$ tokens. This indicates that the codec primitives contain a limited amount of signal per fused P-frame, and that $8$ tokens are sufficient to encode the relevant temporal variations. For this reason, we adopt $8$ $\Delta$-tokens per P-frame as the default configuration throughout, balancing accuracy and token efficiency.

\begin{table}[h]
\centering 
\scriptsize
\caption{\textbf{Number of $\Delta$-tokens per P-frame.} We train several versions of our model with a different number of $\Delta$-tokens per P-frame. Performance significantly increases when going from aggressive compression ($2$ or $4$ tokens) up to $8$, which is the value we used for the results in the main paper. Going to $16$ further improves performance, but with diminishing returns that do not justify the increased token cost.} 
\label{tab:delta_token_comparison} 
\scriptsize
    \begin{tabular}{cll} 
        \toprule 
        \textbf{$\Delta$-tokens per P-Frame} & \textbf{PerceptionTest}  & \textbf{NextQA} \\
        \midrule 
        \multicolumn{3}{l}{\cellcolor[HTML]{EEEEEE}{ \textit{1 Keyframe per GOP}}} \\
        $2$ & $63.26$ & $75.04$ \\
        $4$ & $65.68\smash{\raisebox{+1.9ex}{\scriptsize\textcolor{ForestGreen}{+2.42}}}$ & $76.08\smash{\raisebox{+1.9ex}{\scriptsize\textcolor{ForestGreen}{+1.04}}}$ \\
        $8$ & $67.33\smash{\raisebox{+1.9ex}{\scriptsize\textcolor{ForestGreen}{+1.65}}}$ & $77.37\smash{\raisebox{+1.9ex}{\scriptsize\textcolor{ForestGreen}{+1.29}}}$  \\
        $16$ & $67.69\smash{\raisebox{+1.9ex}{\scriptsize\textcolor{ForestGreen}{+0.36}}}$ & $77.89\smash{\raisebox{+1.9ex}{\scriptsize\textcolor{ForestGreen}{+0.52}}}$ \\
        \arrayrulecolor{black}
        \bottomrule 
    \end{tabular}
\end{table}

\subsection{Two-Stage Training}
\label{subsec:two-stage}
We ablate the contribution of each component in our two-stage training pipeline: (i) pre-training the $\Delta$-Encoder to align codec primitives with the RGB embedding space, and (ii) end-to-end fine-tuning of the full VideoLM. The results are shown in Tab.~\ref{tab:effect_pretrain}. Fine-tuning the VideoLM without any $\Delta$-encoder pre-training yields reasonable performance, though noticeably worse than the full two-stage approach. In this one-stage setting, the model must simultaneously learn motion–residual interpretation, feature alignment, and multimodal reasoning, which slows convergence and leads to weaker temporal understanding, particularly on PerceptionTest. The best results are achieved when both stages are used. Pre-training ensures that $\Delta$-tokens inhabit a well-structured embedding space, while fine-tuning teaches the LLM how to fuse the I- and P-frame tokens together. This combination consistently provides the strongest performance across both datasets. While two-stage training is clearly advantageous in our data regime, we note that the benefit may diminish when training on significantly larger datasets. In such high-data settings, one-stage training may gradually compensate for the missing pre-training at the cost of more compute.

\begin{table}
\centering 
\scriptsize 
\caption{\textbf{Two-stage training.} We attempt to directly train the $\Delta$-encoder together with the LLM without our initial pre-training scheme. This yields significantly lower performance than the two stage setup proposed in the main paper: Stage 1 -- Pre-train the $\Delta$-encoder and Stage 2 -- Fine-tune pre-trained $\Delta$-encoder \& pre-trained LLM together.} 
\label{tab:effect_pretrain} 
\begin{tabular}{ccccc} 
    \toprule 
    \textbf{Pre-train $\Delta$-Encoder} & \textbf{Fine-tune LLM} & \textbf{PerceptionTest} & \textbf{NextQA} \\
    \midrule 
    \multicolumn{4}{l}{\cellcolor[HTML]{EEEEEE}{ 
    \textit{1 Keyframe per GOP}}} \\
    - & \checkmark & $63.45$ & $74.56$ \\
    \checkmark & \checkmark & $\bf 67.33$ & $\bf 77.37$ \\

    \arrayrulecolor{black}
    \bottomrule 
\end{tabular}
\end{table}

\subsection{Are $\Delta$-tokens used by the VideoLM?}
\label{subsec:delta-tok-used}

A natural question is whether the VideoLM actually \emph{uses} the $\Delta$-tokens during inference, rather than ignoring them and relying solely on sparse keyframes. To probe this, we evaluate with $1$ keyframe $+$ $7$ P-frames per GOP but \textit{zero out} all P-frame $\Delta$-tokens at inference time, preserving temporal structure while removing all codec
information. As shown in Tab.~\ref{tab:use_delta_tokens}, performance drops
substantially, confirming that the model actively leverages $\Delta$-tokens for temporal reasoning rather than merely interpolating between sparse I-frames.

\begin{table}[h]
\centering
\scriptsize
\caption{\textbf{Ablation on the use of $\Delta$-tokens.} We compare our method with a version where we set all $\Delta$-tokens to zero, effectively not providing any useful information to the VideoLM. Zeroing out P-frame $\Delta$-tokens leads to a clear performance degradation, confirming that the VideoLM actively attends to and utilizes these tokens.}
\label{tab:use_delta_tokens}
\setlength{\tabcolsep}{6pt}
\begin{tabular}{lcc}
\toprule
\textbf{Sampling Strategy} & \textbf{PerceptionTest} & \textbf{NextQA} \\
\midrule
\multicolumn{3}{l}{\cellcolor[HTML]{EEEEEE}{ 
\textit{1 Keyframe per GOP}}} \\
+ $\Delta$-tokens = $0$ & $64.41$ & $74.21$ \\
+ $\Delta$-tokens & $\bf 67.33$ & $\bf 77.37$ \\
\bottomrule
\end{tabular}
\end{table}

\subsection{Benefits of Codec Primitives}
\label{subsec:controlled-ablation}
In Sec.~\ref{sec:main_results}, we compare our method against LLaVA-Video-7B at matched keyframe counts. To disentangle the contribution of $\Delta$-tokens at inference from the effect of training with codec primitives, we evaluate our
model using only I-frames (Tab.~\ref{tab:controlled_experiment}). Two findings emerge. First, our model with I-frames only already outperforms LLaVA-Video-7B at the same keyframe density, indicating that fine-tuning with interleaved codec primitives teaches the model stronger temporal representations that transfer even when $\Delta$-tokens are absent at inference. Second, adding $\Delta$-tokens on top consistently improves accuracy across all benchmarks, confirming that codec primitives provide complementary temporal signals beyond what the model
internalizes during training. Together, these results show that codec-aware training and $\Delta$-token inference are both independently beneficial and additive.

\begin{table}[h]
\centering
\scriptsize
\caption{\textbf{Controlled ablation: codec-aware training with and without $\Delta$-tokens.} We evaluate our model using only I-frames at each keyframe density, removing $\Delta$-tokens at inference to isolate their contribution. Even without $\Delta$-tokens, our model outperforms LLaVA-Video at matched keyframe counts, showing that fine-tuning with codec primitives alone strengthens temporal reasoning. Adding $\Delta$-tokens consistently improves accuracy further, confirming they provide complementary temporal signals at inference.}
\label{tab:controlled_experiment}
\setlength{\tabcolsep}{4pt}
\begin{tabular}{c c c c c c}
\toprule
 \multirow{2.5}{*}{\textbf{Sampling}} &
\textbf{PerceptionTest} &
\textbf{NextQA} &
\textbf{ActNet-QA} &
\textbf{TempCompass} &
\textbf{TOMATO} \\
\cmidrule(lr){2-2} \cmidrule(lr){3-3} \cmidrule(lr){4-4} \cmidrule(lr){5-5} \cmidrule(lr){6-6}
 & val & mc & test & test-mc & test \\
\midrule
~~1 keyframe / GOP
 & $63.1$ & $78.0$ & $61.8$ & $60.3$ & $22.4$ \\
 
\textcolor{blue}{+ 7 P-frames / GOP} 
 & $65.5$ & $78.3$ & $62.3$ & $62.4$ & $26.2$ \\
 \arrayrulecolor{black!70}\hdashline[0.5pt/2pt]\arrayrulecolor{black}
 
~2 keyframes / GOP
 & $66.2$ & $79.4$ & $62.5$ & $64.5$ & $26.5$ \\
 
\textcolor{blue}{+ 6 P-frames / GOP}
 & $\nd 68.7$ & $\rd 80.4$ & $\rd 63.6$ & $\rd 65.6$ & $\nd 27.1$ \\
 \arrayrulecolor{black!70}\hdashline[0.5pt/2pt]\arrayrulecolor{black}
 
~4 keyframes / GOP
 & $\rd 68.1$ & $\nd 80.8$ & $\nd 63.7$ & $\nd 66.2$ & $\rd 26.9$ \\
 
\textcolor{blue}{+ 4 P-frames / GOP}
 & $\fs 70.3$ & $\fs 82.1$ & $\fs 64.8$ & $\fs 68.9$ & $\fs 28.4$ \\
 
\bottomrule
\end{tabular}
\end{table}

\subsection{Scaling to Higher Frame Rates}
\label{subsec:higher-fps}
A key advantage of our tokenization is that P-frames are represented with only
$N{=}8$ $\Delta$-tokens, compared to $M{=}196$ tokens for each I-frame, enabling higher effective frame rates under a constrained token budget. To study this trade-off, we increase the effective FPS from $1$ to $3$ by
reducing the P-frame fusion window $s$ from $30$ to $10$ frames, thereby encoding finer-grained temporal changes at increasing token cost. We evaluate on TempCompass~\cite{liu2024tempcompass} and
MVBench~\cite{li2024mvbench}, as both benchmarks are sensitive to temporal coverage. As shown in Tab.~\ref{tab:higher_fps}, both benchmarks improve
from $1$ to $2$ FPS, with MVBench gaining over $1.8\%$ and TempCompass nearly $2\%$. However, performance slightly decreases at $3$ FPS on both benchmarks. We note that our model is trained exclusively with $s{=}30$ ($1$ FPS); the
performance drop at higher frame rates likely reflects a train-test mismatch rather than a fundamental limitation, and could be addressed by training with
varied fusion windows. Importantly, these improvements show that our framework enables a practical and flexible trade-off between temporal resolution and token efficiency that would be prohibitive with RGB representations.

\begin{table}[h]
	\centering
    \scriptsize
    \caption{\textbf{Effect of higher frame rates.} We increase the effective FPS by reducing the P-frame fusion window $s$, encoding finer temporal changes at modest token cost. All configurations use $1$ keyframe per GOP. Each I-frame contributes $196{+}14$ tokens (vision $+$ newline), and each P-frame group contributes $8{+}2$ tokens ($\Delta$-tokens $+$ newline), giving $(196{+}14) \times K + (8{+}2) \times P$ tokens per GOP, where $K$ and $P$ are the number of I-frames and P-frame groups respectively.}
    \label{tab:higher_fps}
       \begin{tabular}{cccccc}
            \toprule
            \textbf{FPS} & \textbf{Fusion $s$} & \textbf{P-frames / GOP} & \textbf{Num. Tokens / GOP} & \textbf{TempCompass} & \textbf{MVBench} \\
            \midrule
            $1$ & $30$ & $4$ & $880$ & $66.89$ & $59.98$ \\
            $2$ & $15$ & $8$ & $1760$ & $\fs 68.86$ & $\fs 61.87$ \\
            $3$ & $10$ & $12$ & $2640$ & $\nd 68.73$ & $\nd 61.67$ \\
            \bottomrule
        \end{tabular}
\end{table}

\subsection{Next-Frame Retrieval using $\Delta$-Encoder}
\label{subsec:next_frame_retrieve}
To assess the representational quality of our compressed-domain features, we evaluate next-frame retrieval at $1$~FPS on PerceptionTest~\cite{patraucean2023perception}.
Formally, the task can be defined as: given a query frame $I^{(t-1)}$ at time $t-1$, the task is to identify its true successor frame $I^{(t)}$ at time $t$ from a database containing all frames from the same video (except itself $I^{(t - 1)}$).
As a baseline, SigLIP~\cite{siglip} processes the raw RGB frame $I^{(t-1)}$, whereas our model uses $I^{(t-1)}$ together with the current motion vectors $\tau^{(t)}$ and residuals $\delta^{(t)}$ to produce $N{=}8$ $\Delta$-tokens per P-frame via the $\Delta$-Encoder and lightweight transformer branches used during pre-training.
For this experiment, we use the intermediate checkpoint after pre-training, as it is compatible with the ``reference'' and ``warped'' transformers that are used to transform the features of the previous frame.
The results are shown in Tab.~\ref{tab:retrieval}.
Our method outperforms the baseline by a large margin across the board.
When retrieving a single frame, the relatively low absolute performance is expected, as subsequent frames are heavily similar even at $1$ FPS, but our method successfully leverages the codec primitives to significantly improve the performance. Furthermore, our method achieves almost perfect performance at $5$ frames, with its very high recall of $94.86\%$.
These strong improvements over SigLIP confirm that the $\Delta$-Encoder preserves semantically meaningful motion and appearance cues that are critical for retrieval.
This experiment also highlights that codec primitives contain rich temporal information that is often lost when sampling sparse keyframes.

\begin{table}[t]
\centering
\scriptsize
\caption{{\bf Retrieval performance at 1 FPS.} We evaluate the task of next-frame retrieval. The baseline uses the SigLIP embedding of the current frame, while we use the current frame along with the motion vectors and residuals to next frame, as processed by our $\Delta$-encoder and pre-training transformers. The significantly better performance of our method shows that the information of the codec primitives is correctly compressed in the $\Delta$-tokens.}
\label{tab:retrieval}
\setlength{\tabcolsep}{6pt}
\begin{tabular}{l cccc}
\toprule
\multirow{2}{*}{\textit{@1 FPS}} &
\multicolumn{3}{c}{\textbf{Recall}} \\
\cmidrule(lr){2-4}
 & \textbf{@1} & \textbf{@2} & \textbf{@5} \\
\midrule
SigLIP & $11.12$ & $47.11$ & $78.65$ \\
\textit{Ours - $\Delta$-Encoder} & $\textbf{30.09}$ & $\textbf{77.15}$ & $\textbf{94.86}$ \\
\bottomrule
\end{tabular}
\end{table}

\begin{figure*}
    \centering
    \includegraphics[width=\linewidth]{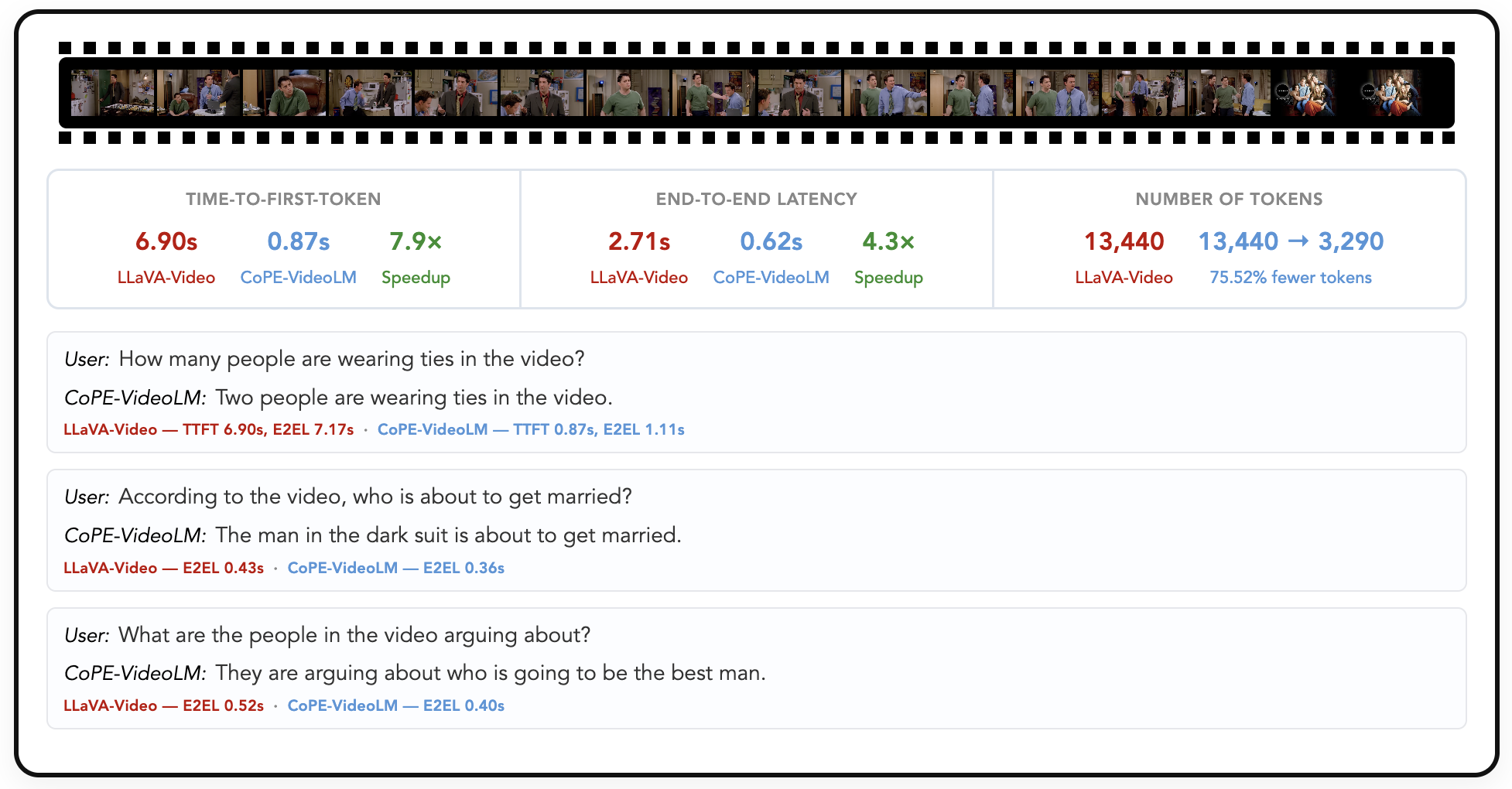}
    \caption{
    \textbf{Qualitative Results.} Multi-turn QA on a TV show clip requiring understanding of character relationships and narrative events. Our method achieves $8.2\times$ faster TTFT and $76\%$ fewer tokens.}
    \label{fig:quals_1}
\end{figure*}

\section{Qualitative Results}
\label{sec:quals_results}
\begin{figure*}
    \centering
    \includegraphics[width=\linewidth]{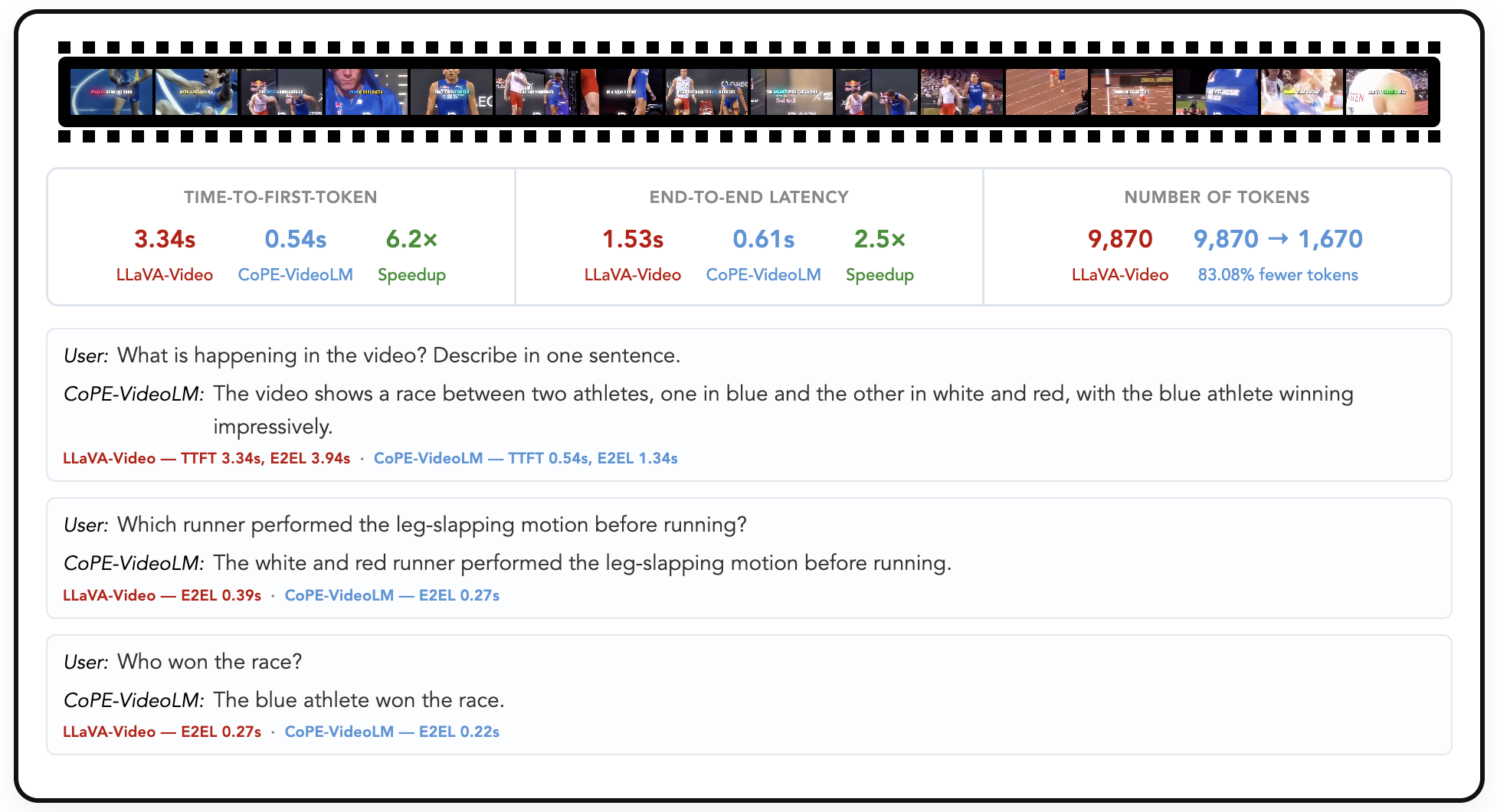}\\[0.5em]
    \includegraphics[width=\linewidth]{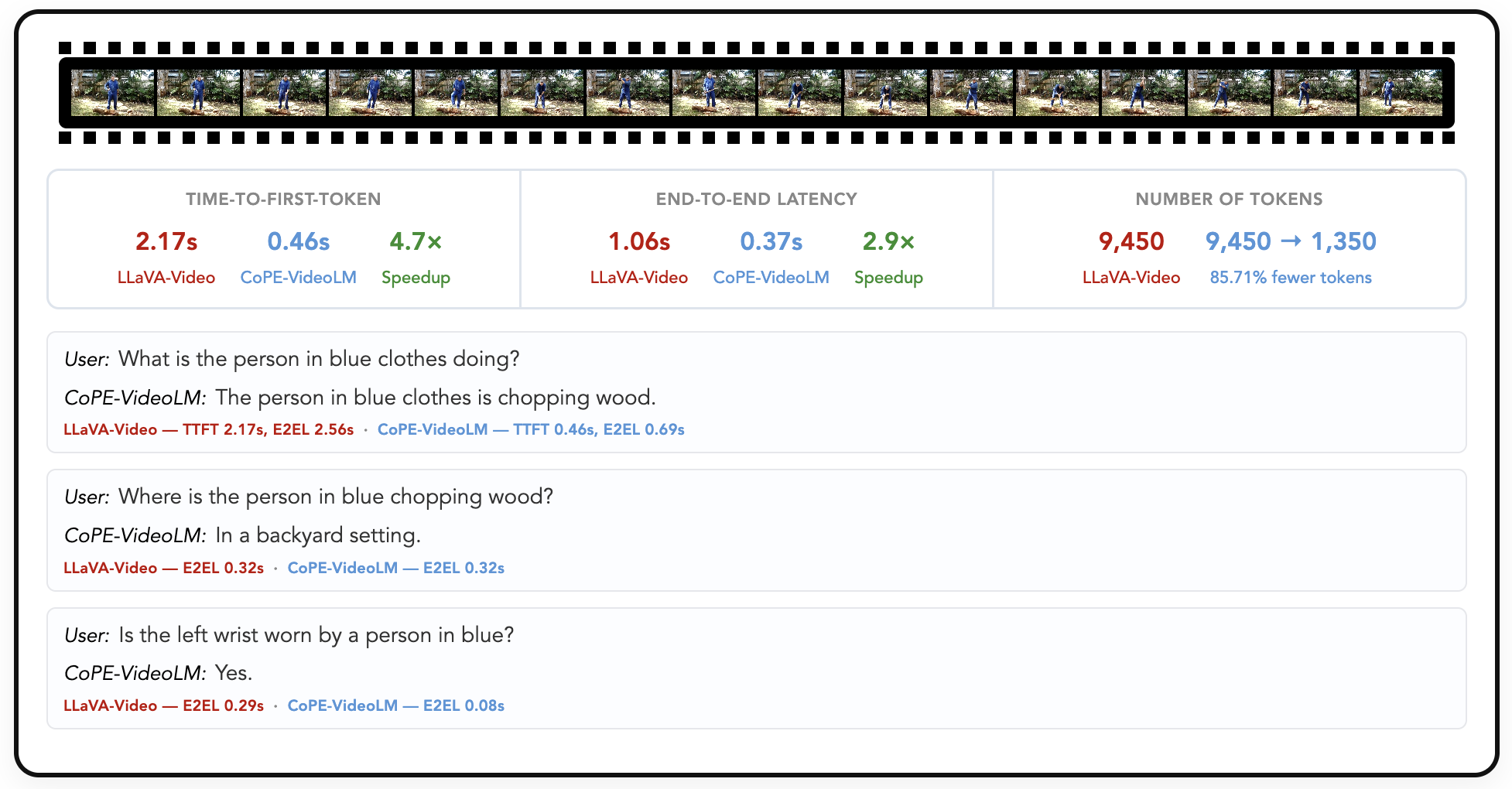}
    \caption{
    \textbf{Qualitative Results.} An athletics race requiring fine-grained action understanding and an outdoor activity video involving spatial and action reasoning. Our method achieves $4.2{-}6.2\times$ faster TTFT and $84{-}86\%$ fewer tokens while producing correct, fine-grained answers.
    }
    \label{fig:quals_2}
\end{figure*}

\begin{figure*}
    \centering
    \includegraphics[width=0.95\linewidth]{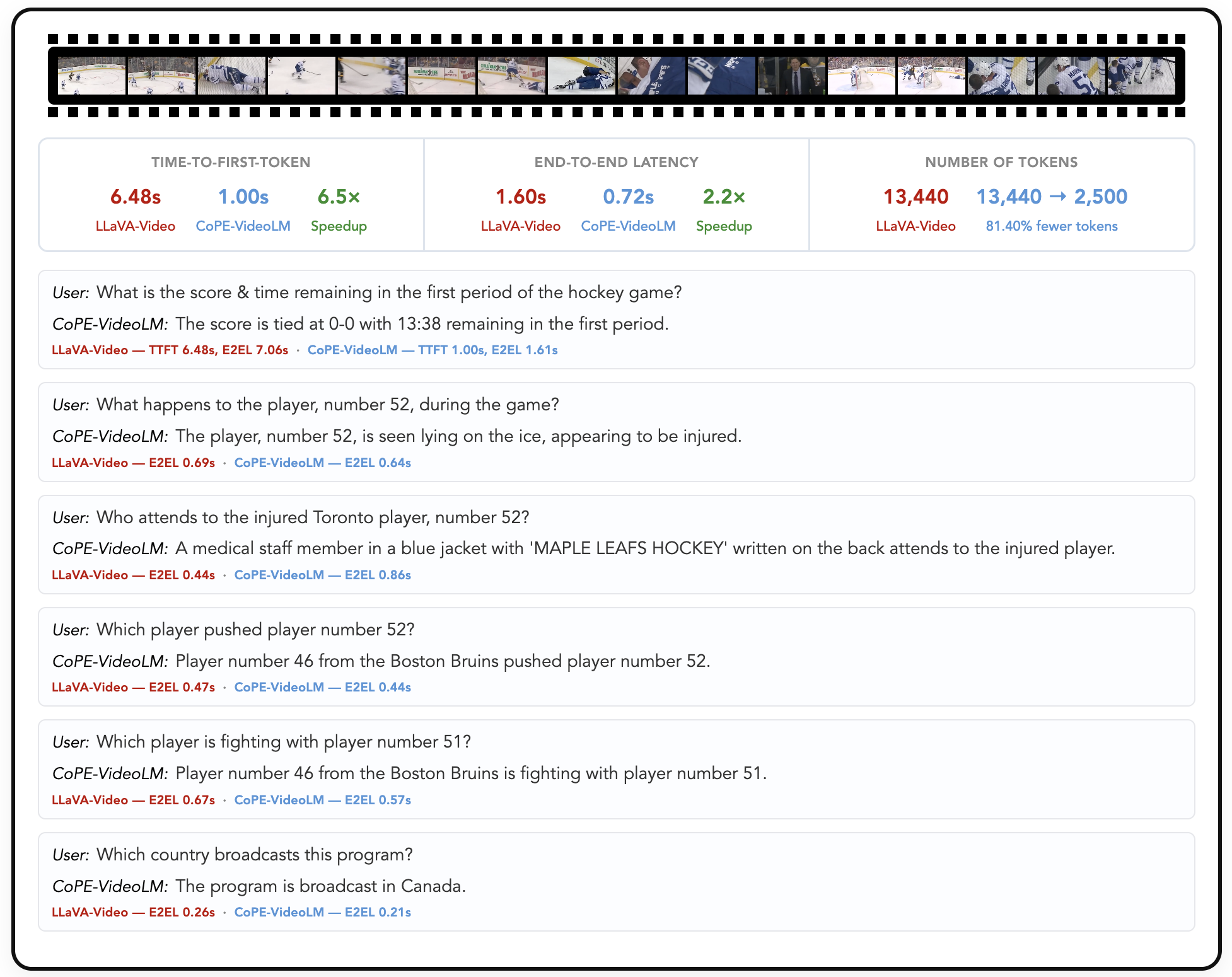}\\[0.5em]
    \includegraphics[width=\linewidth]{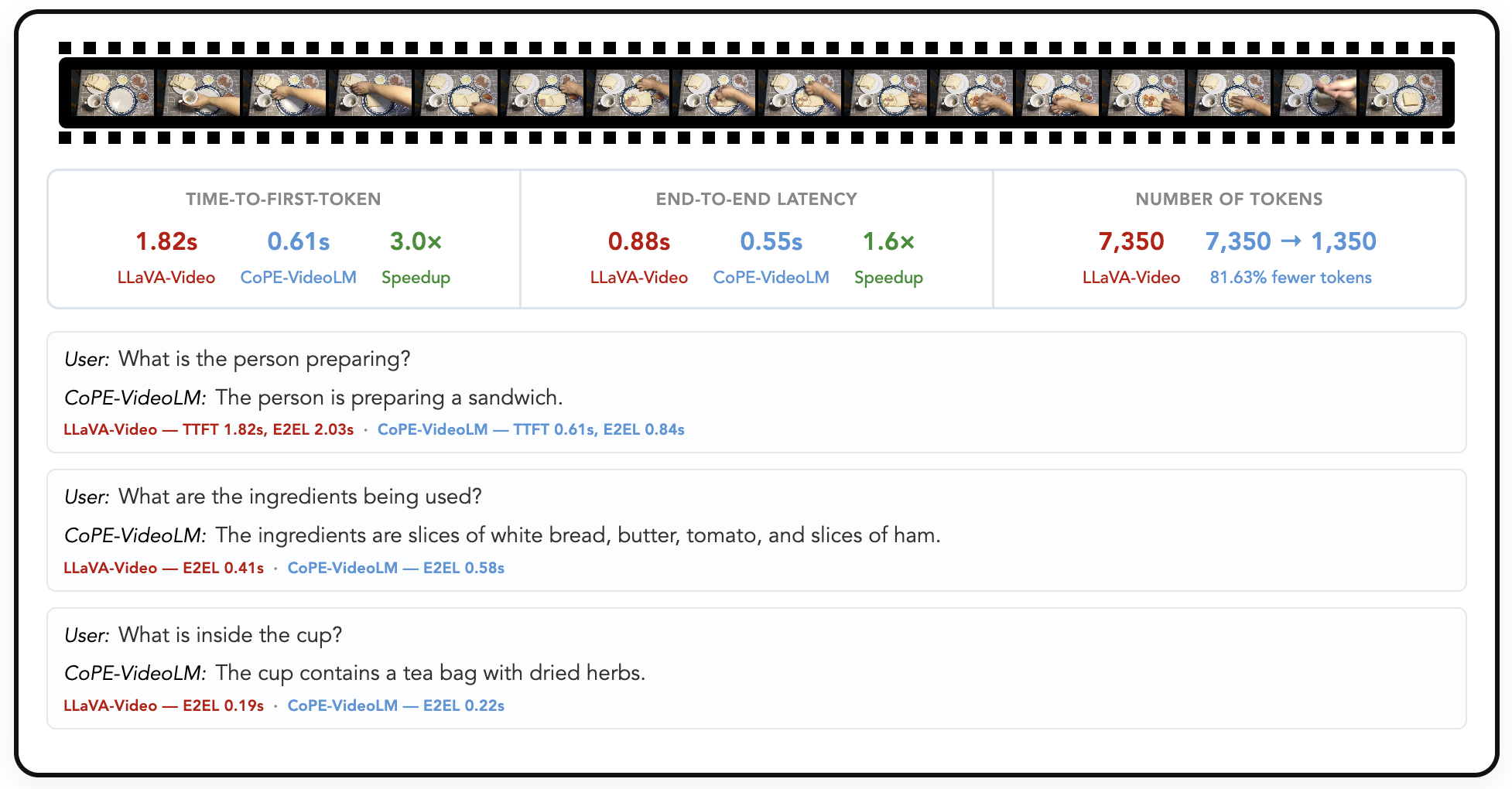}
    \caption{
    \textbf{Qualitative Results.} A hockey game involving multi-step temporal reasoning (injury, player identification, causal events) and, a cooking video with questions about actions, ingredients, and fine-grained object details. Our method provides accurate responses at $2.4{-}3.9\times$ faster TTFT and $82\%$ fewer tokens.
    }
    \label{fig:quals_3}
\end{figure*}

We present qualitative comparisons between CoPE-VideoLM and LLaVA-Video-7B across several videos spanning diverse scenarios: sports with fast motion (hockey, athletics), indoor scenes (TV show, cooking), and outdoor activities (wood chopping). For each example, we report TTFT, E2EL, and token counts for both methods, alongside CoPE-VideoLM's multi-turn QA responses. Note that from the second turn onward, both methods benefit from KV-cache reuse, which narrows the absolute latency gap. However, our method still maintains a consistent relative speedup across all turns due to the shorter cached sequence length from fewer tokens. As shown in Figs.~\ref{fig:quals_1}--\ref{fig:quals_3}, CoPE-VideoLM produces correct, fine-grained answers while achieving consistent speedups at substantially reduced token counts.

% \clearpage
{
    \small
    \bibliographystyle{unsrtnat}
    \bibliography{main}
}

\end{document}